\documentclass[preprint,journal]{vgtc}       

\usepackage{soul}

\ifpdf
  \pdfoutput=1\relax                   
  \usepackage{graphicx}                
  \DeclareGraphicsExtensions{.pdf,.png,.jpg,.jpeg} 
\else
  \usepackage{graphicx}                
  \DeclareGraphicsExtensions{.eps}     
\fi%

\graphicspath{{figures/}{pictures/}{s/}{./}} 

\usepackage{microtype}                 
\PassOptionsToPackage{warn}{textcomp}  
\usepackage{textcomp}                  
\usepackage{mathptmx}                  
\usepackage{amssymb}
\usepackage{times}                     
\usepackage{cite}                      
\usepackage{tabu}                      
\usepackage{booktabs}                  
\usepackage{algorithm}
\usepackage[noend]{algpseudocode}
\usepackage[caption=false]{subfig}
\usepackage{amsmath}
\usepackage{fixmath}

\usepackage[caption=false]{subfig}
\makeatletter
\def\BState{\State\hskip-\ALG@thistlm}
\makeatother

\newcommand{\myparagraph}[1]{\mbox{\ } \newline \noindent \textbf{#1}}
\renewcommand{\paragraph}[1]{\myparagraph{#1}}

\usepackage{color}
\definecolor{mscolor}{rgb}{0.1,0.8,0.3}
\definecolor{mccolor}{rgb}{0.2,0.0,0.4}
\definecolor{mygreencolor}{rgb}{0.3,0.8,0.3}
\definecolor{purple}{rgb}{0.65,0,0.65}

\newcommand{\yh}[1]{{\color{black} #1}}

\usepackage{multirow}
\newcommand*{\rot}[2]{
	\rotatebox[origin=l]{#1}{#2}
}

\newcommand{\RR}{\mathbb{R}}
\renewcommand{\vec}[1]{\mathbold{#1}}

\newcommand{\rev}[1]{\textcolor{black}{#1}}

\onlineid{1330}

\vgtccategory{Research}
\vgtcpapertype{algorithm/technique}

\title{Implicit Multidimensional Projection of Local Subspaces}

\author{Rongzheng Bian, Yumeng Xue, Liang Zhou, Jian Zhang, Baoquan Chen, Daniel Weiskopf, and Yunhai Wang}
\authorfooter{
\item
R. Bian, Y. Xue and, Y. Wang are with Shandong University, Qingdao. Email: \{bianrongzheng, xym6336, cloudseawang\}@gmail.com.
\item
B. Chen is with Peking University. E-mail: baoquan.chen@gmail.com.
\item
J. Zhang is with CNIC, CAS. E-mail: zhangjian@sccas.cn.
\item
L. Zhou is with University of Utah. E-mail: lzhou@sci.utah.edu
\item
D. Weiskopf is with University of Stuttgart. E-mail: weiskopf@visus.uni-stuttgart.de
\item
Y. Wang and L. Zhou are joint corresponding authors.

}

\shortauthortitle{Biv \MakeLowercase{\textit{et al.}}: Global Illumination for Fun and Profit}

\abstract{We propose a visualization method to understand the effect of multidimensional projection on local subspaces, using implicit function differentiation.
	Here, we understand the local subspace as the multidimensional local neighborhood of data points. 
	Existing methods focus on the projection of multidimensional data points, and the neighborhood information is ignored. 
	Our method is able to analyze the shape and directional information of the local subspace to gain more insights into the global structure of the data through the perception of local structures.
	Local subspaces are fitted by multidimensional ellipses that are spanned by basis vectors. 
	An accurate and efficient vector transformation method is proposed based on analytical differentiation of multidimensional projections formulated as implicit functions.	
	The results are visualized as glyphs and analyzed using a full set of specifically-designed interactions supported in our efficient web-based visualization tool.
	The usefulness of our method is demonstrated using various multi- and high-dimensional benchmark datasets.
	Our implicit differentiation vector transformation is evaluated through numerical comparisons; the overall method is evaluated through exploration examples and use cases.
}


\teaser{
  \centering

  \includegraphics[width=\linewidth]{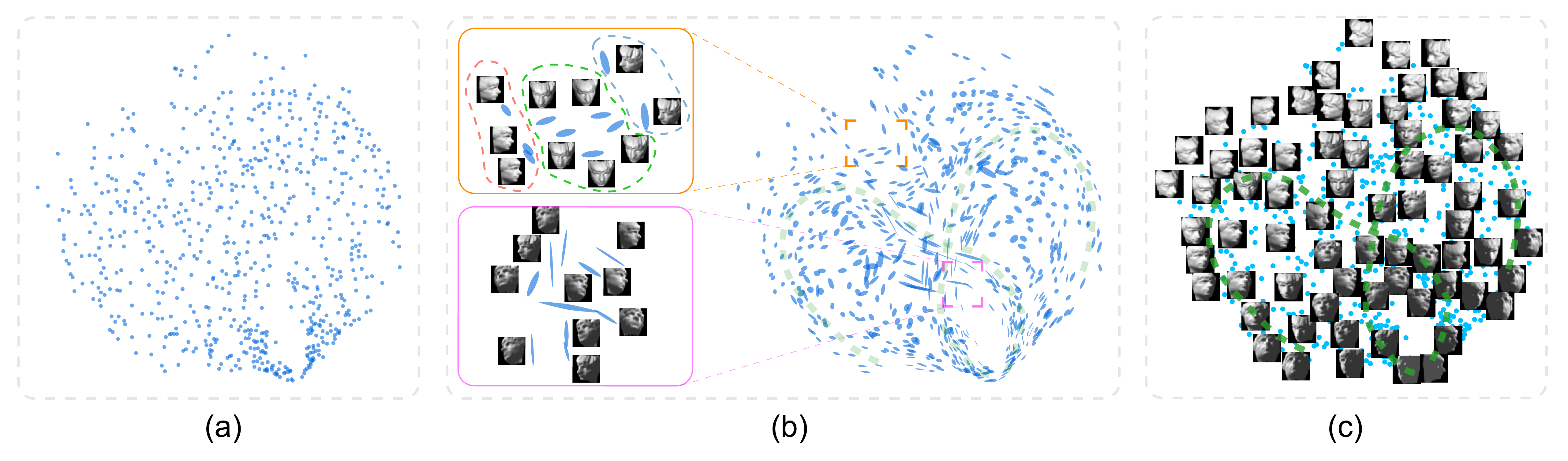}
  \vspace{-9mm}
\caption{The face dataset is projected to 2D using nonlinear MDS.
(a)~Traditional visualization just shows the projected data points; here, no obvious correlation can be seen. (b)~With our new implicit local subspace projection method, global trends and local structures (i.e., local linear subspaces around data points in the original high-dimensional space) are visualized through the orientation and shape of glyphs that replace the dots in the 2D plot. Now, two groups of glyphs associated with right- and left-facing faces are clearly separated (near the bottom), and some more of the interesting local patterns are identified in the zoom-ins. (c) Two trends that cross at the center of the plot are highlighted in green in the image view (also in (b), transparent dashed lines in green); such crossing would not be visible without the orientation encoding in our new glyphs.  }
	\label{fig:teaser}
	\vspace{-1mm}
}

\renewcommand{\manuscriptnotetxt}{}

\vgtcinsertpkg

\begin{document}



\firstsection{Introduction}
\maketitle

Multidimensional data analysis has manifold applications in diverse domains such as finance, science, and engineering. It is often conducted by reducing data dimensionality with a dimensionality reduction (DR) technique and then visualizing the reduced data with scatterplots.
By representing data points as visual marks in 2D space, 2D scatterplots are one of the most useful and common approaches~\cite{sedlmair2013empirical} for visual exploration of multidimensional data.
However, depicting the projected data with point-based information alone is oversimplified as only the density distribution can be perceived.
In this paper, we propose a model to visually understand multidimensional local linear subspaces after dimensionality reduction---we refer to local linear subspace as the local linearized neighborhood in the original multidimensional space.

There are many DR techniques that preserve certain structures of the data in low-dimensional space.
Early techniques aim to faithfully represent the data's global structures, for example, principal component analysis (PCA)~\cite{hotelling1933analysis} and multidimensional scaling (MDS)~\cite{Torgerson1965,borg2005modern}, however, they cannot effectively reveal the low-dimensional manifold embedded in high-dimensional data. In contrast, more recent DR methods, e.g., locally linear embedding (LLE)~\cite{roweis2000nonlinear} and t-distributed stochastic neighbor embedding (t-SNE)~\cite{van2008visualizing}, seek to map nearby points in the high-dimensional space to nearby points in the low-dimensional space.
These methods can better preserve local linear structures, however, only encoding data points as positions of visual marks will not reveal such structures, except for the distance between them.
In particular, the traditional point visualization misses any orientation information from the local linear structures.

\autoref{fig:teaser} illustrates this issue for a typical DR example along with our strategy to resolve it.
\autoref{fig:teaser}~(a) shows a traditional dot-based visualization in a scatterplot, where only the density distribution of data points can be perceived. With our new implicit local subspace projection method (\autoref{fig:teaser}~(b)), global trends and local structures are visualized through the orientation and shape of glyphs. With our approach, we can visually separate two different groups of glyphs near the bottom-right of the plot---they are generally associated with right- and left-facing faces (transparent dashed curves in green).
Another example is a global trend highlighted with \yh{two green dashed curves} in the image view in~\autoref{fig:teaser}~(c); now, it is possible to identify two trends that cross at the center of the plot, each is associated with a different smoothly changing camera position.
 \autoref{fig:teaser}~(b) demonstrates that further interesting local pattern can be identified with the glyphs, \yh{where three clusters of glyphs (bounded by dash lines in the orange zoom-in) correspond to the associated images with three distinct face orientations and the crossing pattern (see the purple zoom-in) well reflects the smooth transitions of face directions.}

We propose a model to characterize multidimensional local subspaces with basis vectors anchored at each data point (all in the original space) and transform the basis vectors to a low-dimensional visualization space with an analytical vector transformation technique based on the implicit function theorem.
The basis vectors are extracted using local PCA in the original multidimensional space.
Our vector transformation method takes the DR technique as an implicit function and uses its analytical gradient to compute the accurate projected basis vectors.  In this way, we guarantee that the transformation of the subspace is consistent with the projection of the points in the DR method.
Our implicit function-based vector transformation has the advantage of being efficient and accurate at the same time.
%

To visualize the projected local subspace, we construct a glyph in the form of a closed B-spline curve that captures the deformations introduced by the transformation of the basis vectors.
We reduce the overlap between glyphs by interactively changing their opacities and size, and order them according to glyph area. \yh{We measure the projection quality with two metrics: the loss function of the DR method and trustworthiness~\cite{Espadoto19,Nonato19}, and compute the anisotropy of the local subspace in high dimensional data.}
Combining these measures, we design a set of linked views for interactively examining projection errors and the relationship between them and data attributes.
Based on a web-based implementation, we demonstrate the effectiveness of our framework with case studies.

Our main contributions are as follows:
\begin{itemize}
	\item A model that visualizes local subspaces for multidimensional projections such that global trends in the data can be inferred by the perception of local subspaces;
	\item An analytical vector transformation method based on implicit function differentiation; and
	\item The glyph representation that shows the geometry of transformed basis vectors of local structures in the same 2D domain as a traditional scatterplot.
\end{itemize}

\yh{Since we do not change the basic intuition associated with traditional multidimensional visualization, our approach readily fits in any existing DR visualization pipeline, regardless of whether it uses linear or nonlinear multidimensional projections.  The source code is available for download on GitHub\footnote{{\small \url{https://github.com/VisLabWang/DRImplicitVecXform}}}}.

\renewcommand{\bottomfraction}{0.5}
\renewcommand{\topfraction}{0.9}
 \renewcommand{\textfraction}{0.1}
  \setcounter{bottomnumber}{2}
\setcounter{totalnumber}{5}

\section{Related Work}


Dimensionality reduction is an important research topic in statistics,  data science, and visualization.
A comprehensive survey of DR methods can be found elsewhere~\cite{Lee:2007:NDR:1557216,van2009dimensionality}.
%
One issue of DR techniques is their interpretation. This is a difficult problem~\cite{vellido2012making} addressed in extensive previous work.
A large body of research focuses on providing empirical guidance and design principles for interpreting DR results and assessing their quality.
Some examples include a systematic literature review of the topic~\cite{Sacha:vis17}, metric-based quality assessments of synthetic data~\cite{J.Lee2008Qaon}, and a study of different visual encoding schemes~\cite{sedlmair2013empirical}.

Another branch of research improves the interpretability of DR visualizations by including quality information~\cite{Espadoto19,Nonato19}.
Aupetit~\cite{AUPETIT20071304} encodes Voronoi cells with luminance in the 2D visualization with distortion and uncertainty measurements to show the DR quality.
Similarly, CheckViz~\cite{Lespinats:eurovis11} visualizes distortions in the local mappings with a 2D perceptually uniform color map applied to Voronoi-cell partitionings of the 2D scatterplot after dimensionality reduction.
Seifert et al.~\cite{Seifert:EuroVAST10:013-018} visualize the local stress metric~\cite{chen:jasa09} as a combination of a 2D heat map and a height map, simultaneously showing the stress levels of each datum and its neighboring area.

Interactive tools with brushing-and-linking and carefully designed visual encodings are also useful for understanding DR results.
Embedding Projector~\cite{smilkov2016embedding} allows the user to explore DR data as a 3D scatterplot and shows DR processes by animation; however, it is a general tool for overview and more in-depth analysis is not supported.
\rev{A set of interactive visual analysis methods using various visual encoding are proposed to study the quality of dimensionality reduction methods on large-scale datasets~\cite{MARTINS201426}.}
Stahnke et al.~\cite{Stahnke:vis15} probe multidimensional projections with a set of integrated interaction techniques that allow for the investigation of each data point and a neighborhood as well as additional information, e.g., classes, clusterings, and original dimensions.
Comparative visualization~\cite{cutura2018viscoder} provides results of multiple DR techniques, where the user can assess different behaviors of these techniques through interactive exploration of multidimensional data with linked views.

Relating DR results with original data dimensions is yet another way of interpreting those results.
Axes of original data dimensions can be drawn as 3D biplot~\cite{Gabriel:biometrika71} axes to understand DR results in 3D plots~\cite{coimbra:iv16}.
An interactive framework~\cite{Cavallo:2018:VIF:3173574.3174209} allows the user to visually explore forward (high-to-low) and backward (low-to-high) projections, where original data axes are shown in 2D biplots.
Alternatively, a perturbation-analysis-based method~\cite{faust2019dimreader} aims to understand nonlinear DR methods.
Here, the goal is to recognize how generalized axis lines (visualized as contours) change according to user-specified infinitesimal perturbations; small changes of the data are modeled in the original multidimensional space and their effect on the projected axes in the DR display are computed by automatic differentiation.
%
Our work has a different goal in mind: we want to understand the shape and orientation of the local structure, and, furthermore, global correlations of data points; and we use implicit differentiation to accurately transform basis vectors in the multidimensional local neighborhood.

Features of interest in multidimensional space may live in low-dimensional subspaces~\cite{Vidal_atutorial}.
Therefore, subspace analysis that models multidimensional data by the union of multiple subspaces is a powerful tool for multidimensional data analysis. For example, subspace clustering localizes the search for features in relevant dimensions~\cite{Parsons:2004:SCH:1007730.1007731,Vidal_atutorial}.
Recently, sparse and low-rank subspace clustering methods have been widely used in machine learning, computer vision, and pattern recognition~\cite{Liu:tpami13,Elhamifar:tpami13}.
Related concepts are useful for visualization as well. For example,
automatic subspace searching, grouping, and filtering, followed by interactive analysis allow users to visually explore multidimensional data~\cite{tatu:vast12}.
Alternatively, subspace clustering and animation of dynamic projections can be used for interactive visual exploration of subspaces~\cite{liu:cgf15}.
These methods analyze lower-dimensional subspaces and point-based projections in multidimensional datasets. In contrast, we focus on shape and direction-based analysis of local neighborhoods of the same dimensionality of the data.

\begin{figure*}[htb]
	\centering
	\includegraphics[trim={2cm 2cm 0cm 0cm}, clip, width = \linewidth]{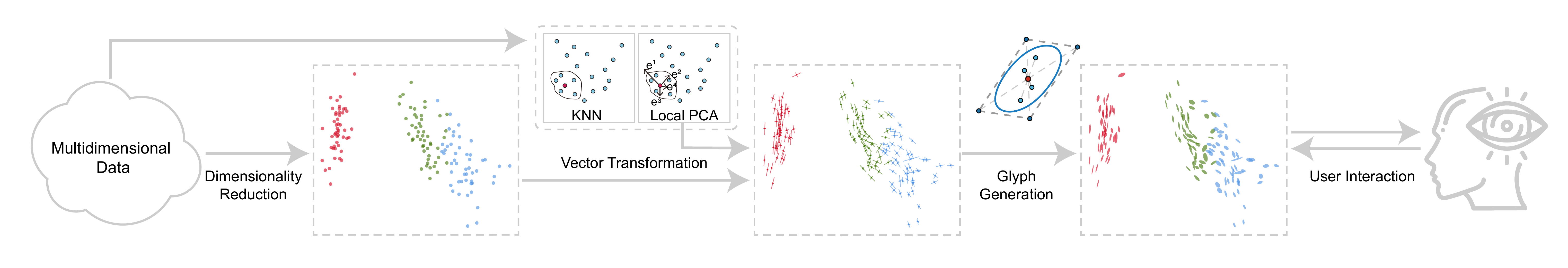}
	\vspace{-5mm}
	\caption{Workflow of our method: Dimensionalty reduction (left) projects points from the original space to the low-dimensional visualization space. We extend this conventional approach by first computing a linear subspace around each data point using local PCA and then projecting the basis vectors of the subspace to the low-dimensional visualization space. These transformed vectors are used to encode the deformation of the projected subspace in glyphs.}
	\label{fig:methodWorkflow}
	\vspace{-6.5mm}
\end{figure*}

Flow-based scatterplots~\cite{chan:vast10} and generalized sensitivity scatterplots~\cite{chan2013generalized} enhance 2D scatterplots with glyphs encoding local trends.
These methods are effective in identifying local relationships between two variables.
More recently, clusters in DR data have been visualized on scatterplots with winglets~\cite{lu2019winglets} that enhance data points with arcs encoding cluster information and uncertainty.
\yh{However, the local patterns visualized by both approaches are not defined by the whole original data. In contrast, improved parallel coordinates plots~\cite{zhou2017indexed,nguyen2017dspcp} can directly visualize local multivariate correlations of the original data. For example, indexed-points parallel coordinates~\cite{zhou2017indexed} represent multidimensional locally fitted planar structures by using p-flat indexed points, allowing for effective pattern recognition through visual clustering. However, such parallel coordinates are effective only for data with a small number of dimensions~\cite{siirtola2009visual}. In contrast, our method is designed for multidimensional projections, can be applied to high-dimensional data, considers multiple vectors to faithfully describe local subspaces, and---most importantly---works consistently with any DR method.}

\yh{Recently, user-centered dimensionality reduction methods that allow for fine tuning of the projections have attracted increasing attention. Most methods~\cite{Paulovich08,Joia11,Fadel15} based on neighborhood reconstruction are designed to support full interactivity and controllability but require user-provided control points. In contrast, our method focuses on the understanding of  projection results with glyph perception and user interaction, and reveals the structure of local subspace around each projection point.
}

\section{Method Overview}
\label{sec:methodoverview}
\begin{figure}[b]
	\centering
	\vspace{-6mm}
	\subfloat[]{\includegraphics[width = 0.33\linewidth]{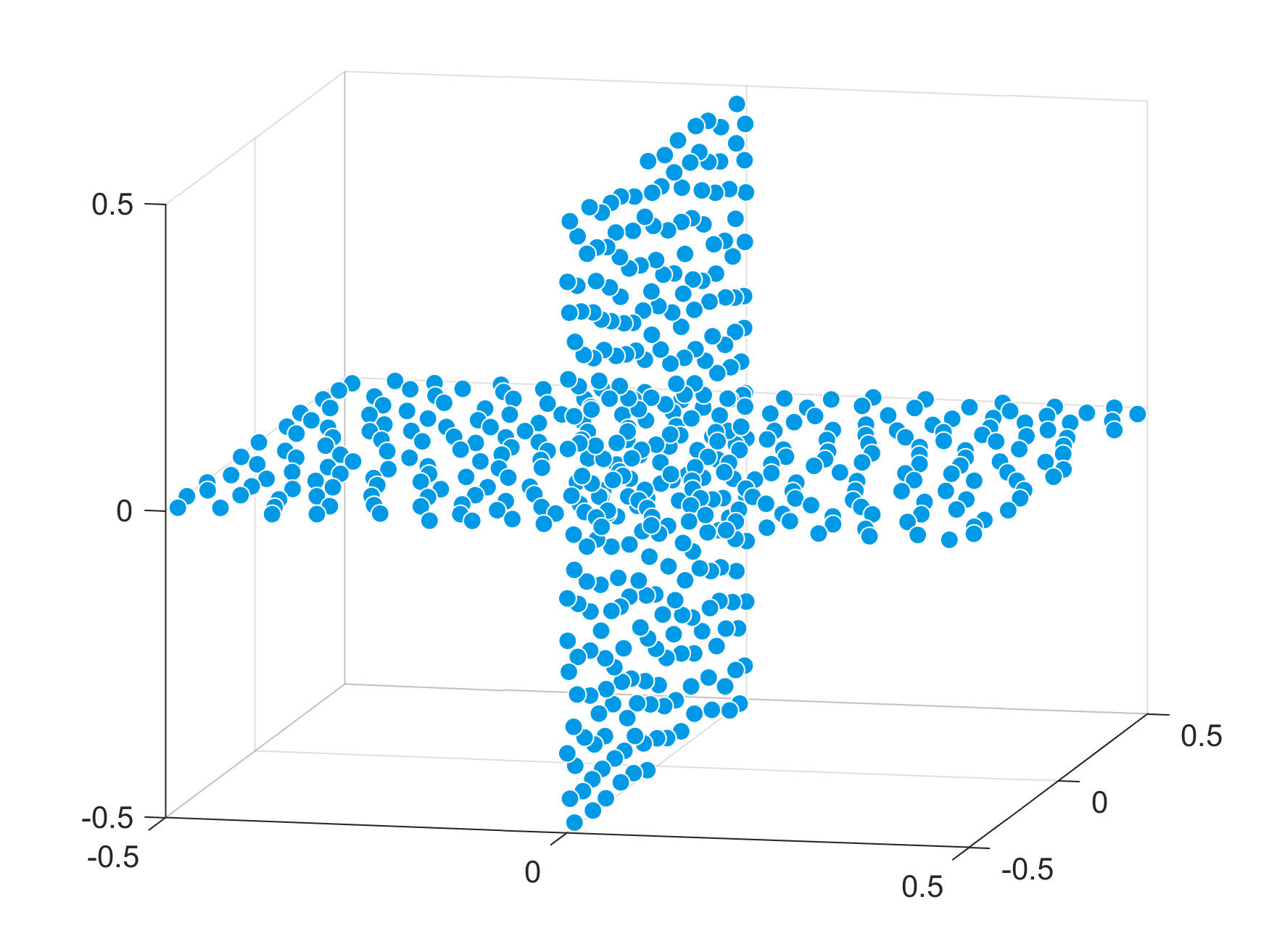}}\hfill
	\subfloat[]{\includegraphics[width = 0.3\linewidth]{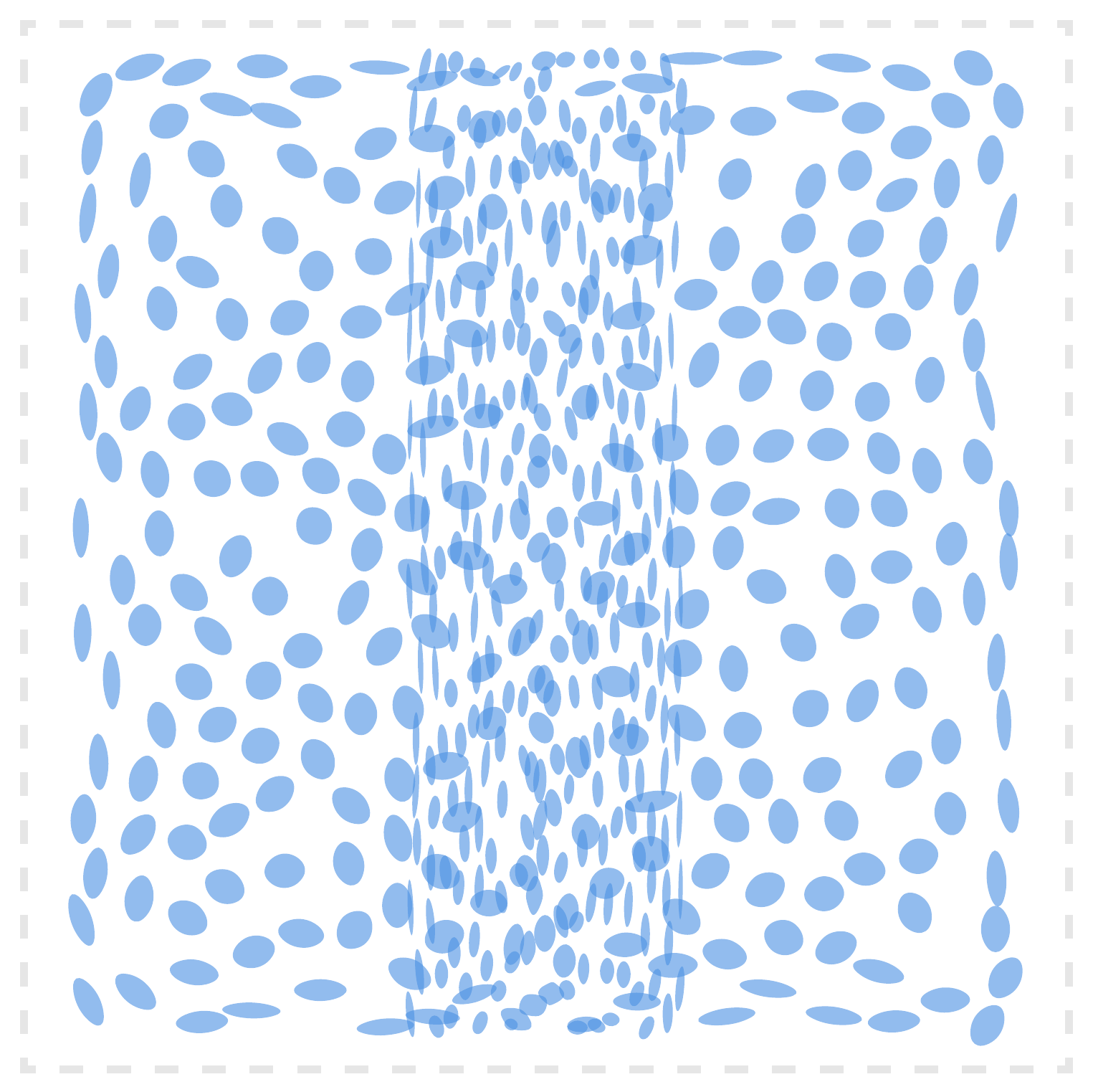}}\hfill
	\subfloat[]{\includegraphics[width = 0.3\linewidth]{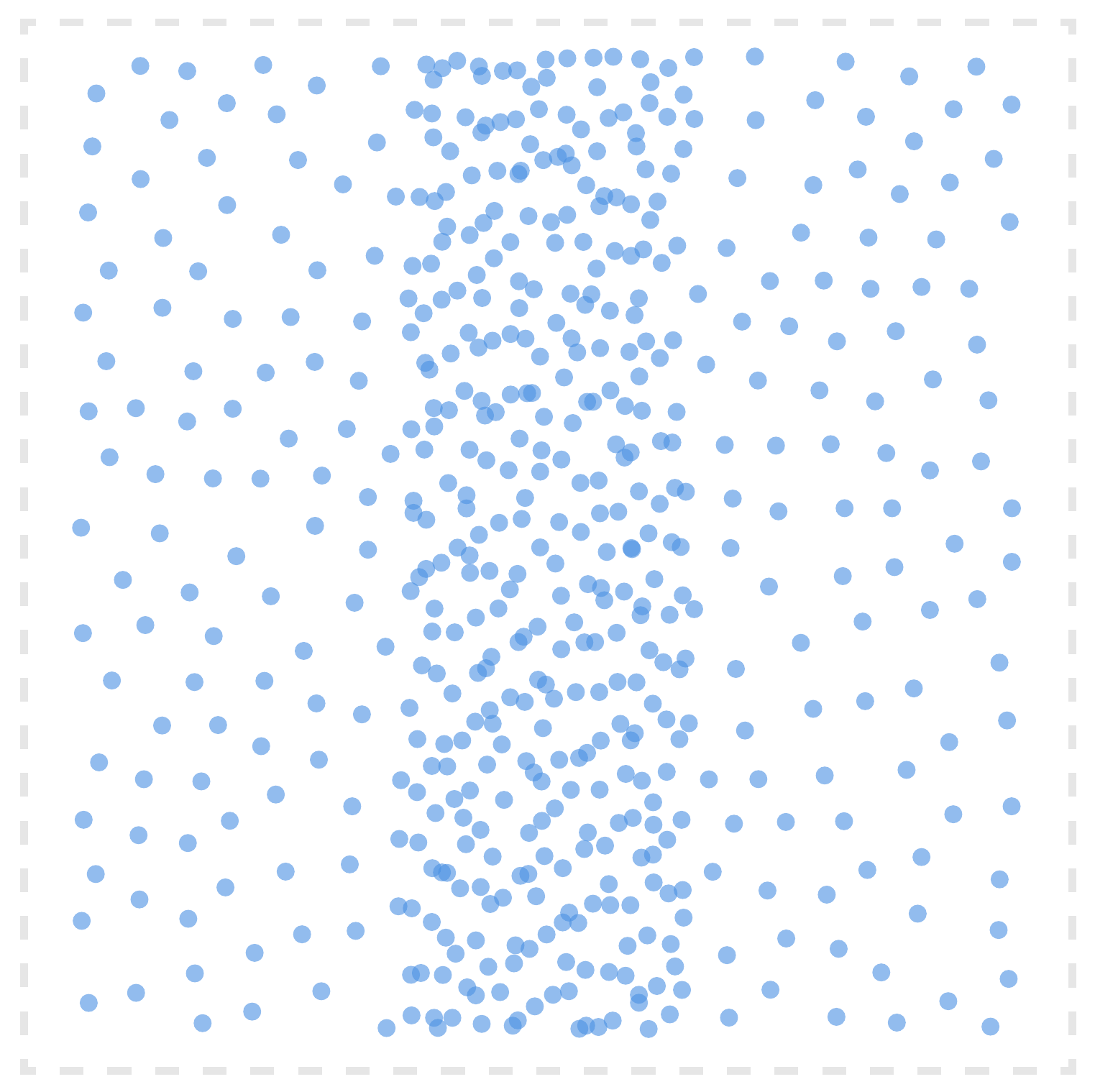}}\\
	\subfloat[]{\includegraphics[width = 0.33\linewidth]{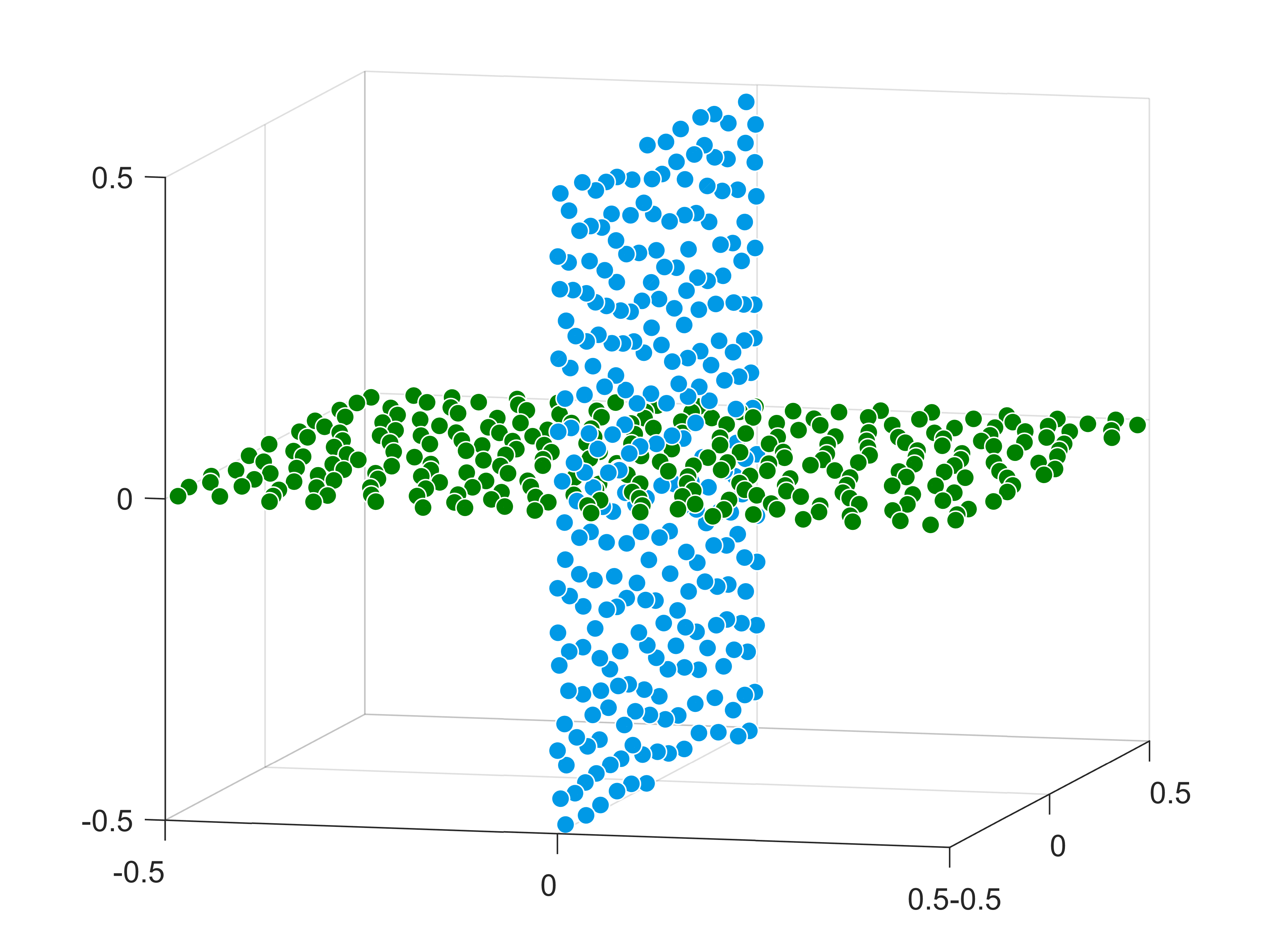}}\hfill
	\subfloat[]{\includegraphics[width = 0.3\linewidth]{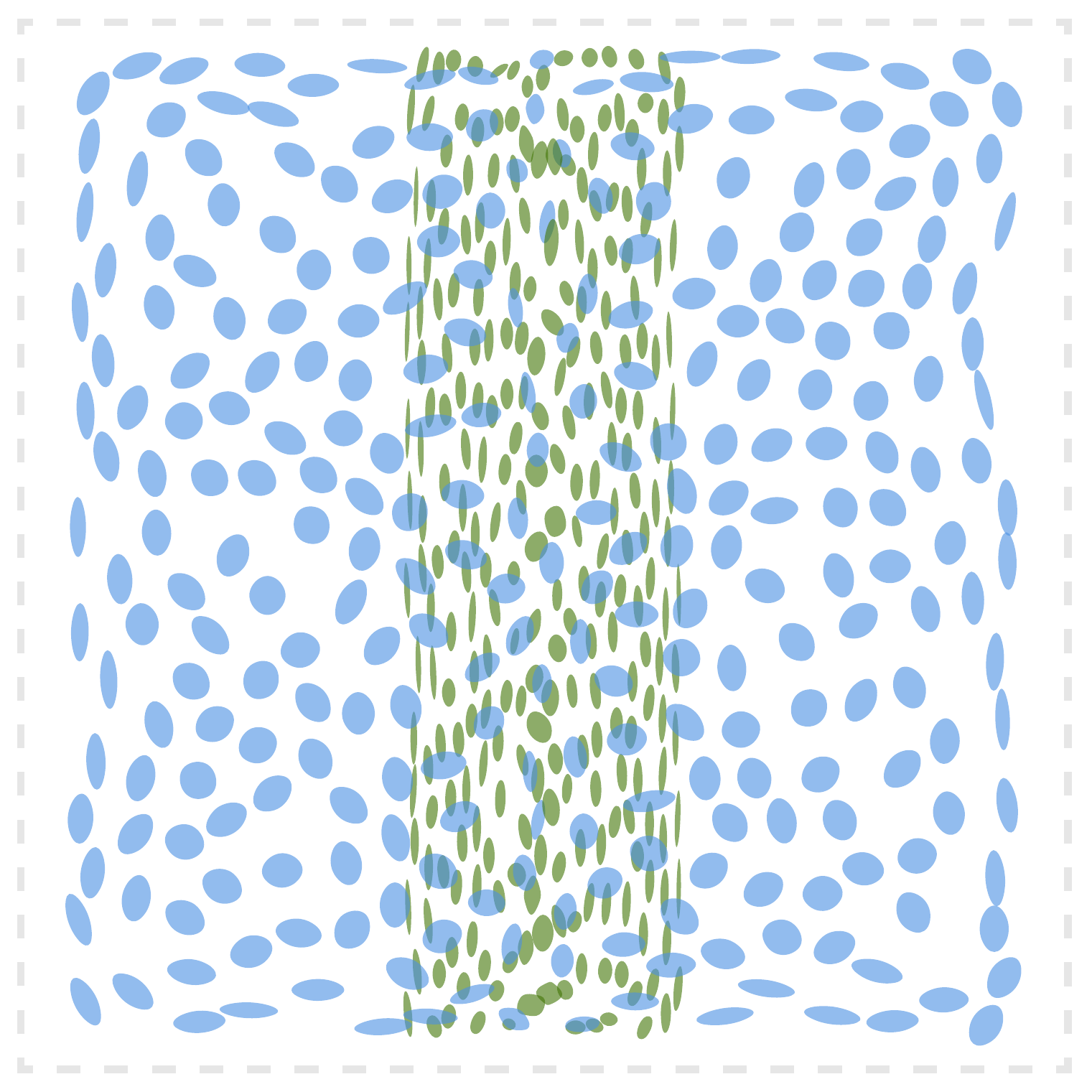}}\hfill
	\subfloat[]{\includegraphics[width = 0.3\linewidth]{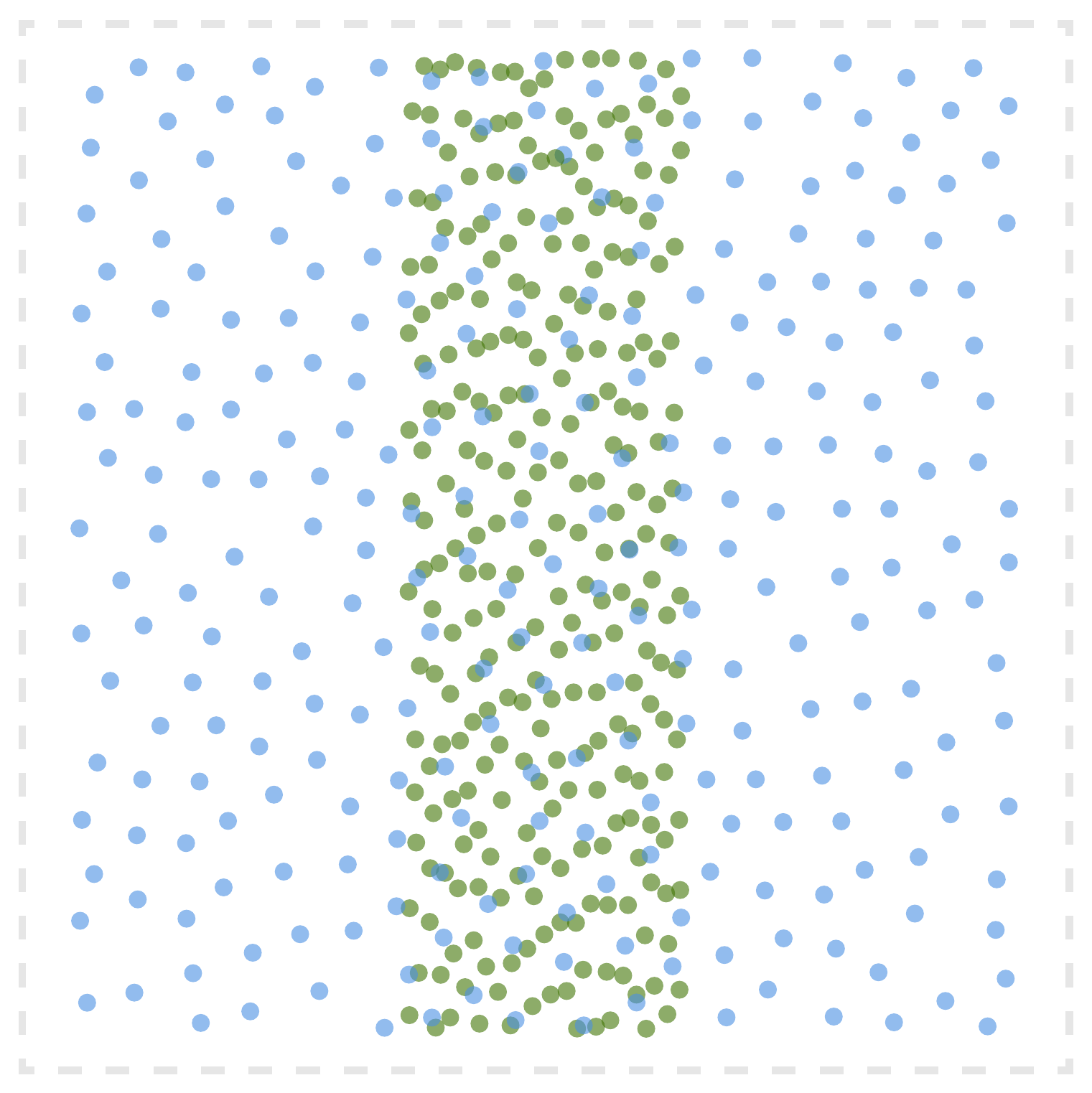}}
	\caption{Two perpendicular planes in 3D (a) and rendered with color-coding (d). The multidimensional projection is visualized using our method (b) and (e), and traditional point-based visualization (c) and (f).}
	\label{fig:plane2}
	\vspace{-3mm}
\end{figure}

Our method augments point-based multidimensional projection with the shape and orientation information \yh{depicting the multidimensional local subspace} around data points.
The global structure of multidimensional datasets can be better understood by including local subspaces.
\autoref{fig:plane2} motivates our approach for a simple example of a synthetic data with two perpendicular planes in 3D space.

The projection is chosen in a way so that the projection direction is not perpendicular to the planes.
Therefore, it is not possible to separate the two planes in the traditional point-based visualization (\autoref{fig:plane2}~(c)) unless we use color to distinguish them (\autoref{fig:plane2}~(f)).
With our method, the two planes can be easily identified---even without color coding---by the shapes and directions of glyphs: thin, elongated, vertical-going glyphs associated with the plane that covers a smaller area after projection; and round glyphs that represent the plane that is better preserved in the projection (\autoref{fig:plane2}~(b)).

Our approach extends traditional multidimensional projection in three ways: (1)  extract the local linear subspace in the original space, (2) project the subspace in a way that is consistent with the DR technique applied to the data points, and (3) visualize the information from the projection of the subspace. The respective workflow of our method is illustrated in~\autoref{fig:methodWorkflow}.

For (1), we identify the local linear structure around each data point in the multidimensional space using the $k$-nearest neighbors (kNN) method.
The local neighborhood of a point is fitted by a multidimensional ellipsoid as we perform PCA of the neighborhood and obtain its eigenvectors and the corresponding eigenvalues.
In step (2), the extracted subspaces are transformed to the 2D visualization space. To this end, we transform the eigenvectors spanning local subspaces using implicit differentiation as explained in Section~\ref{sec:model}. In practice, not all eigenvectors are needed as many of them are associated with small eigenvalues and contribute little to the subsequent computations, i.e., we use local linear subspaces that do not necessarily have the full dimensionality of the original dataset.
Finally (3), data with local subspace information is visualized using 2D ellipse-like glyphs generated from the data points and the transformed vectors as discussed in Section~\ref{sec:visnUI}.




\section{Mathematical Model}
\label{sec:model}

This section first clarifies the terminology and mathematical assumptions that we make. Then, we discuss our way of representing local linear subspaces and the corresponding transformation of vectors. This leads to our approach to using implicit functions to compute the transformation for general nonlinear DR methods.


\subsection{Multidimensional Projection of Points}

We present a general definition of multidimensional projections that will be used as a basis for our extended projection approach. In the following, we use uppercase letters for variables that are related to the original multi-dimensional space, and lowercase letters for quantities in the projection space (i.e., typically the 2D visualization space).

Given a multi-dimensional dataset with the dimensionality $D$,
a point in this space is denoted $P \in \RR^D$. Multidimensional projection takes points at $P$ to corresponding locations $p \in \RR^d$ ($d \leq D$) in the lower-dimensional visualization space: 
\begin{equation}
\pi: \; \RR^D \to \RR^d \, , \quad  P \mapsto p = \pi(P) \;.
\end{equation}
%
Our discussion and derivation is independent of the choice of $\pi$, i.e., we formulate our approach to work with any linear or nonlinear multidimensional projection.

Most nonlinear multidimensional projection techniques employ some complex optimization approaches to arrive at the projection $\pi$. Often, this map is not defined for all $\RR^D$, but only the projection of the input points is computed, i.e., the target locations $p_i = \pi(P_i)$. In contrast, the linear projection $\pi$ is directly accessible. A prominent case is the linear projection by PCA~\cite{hotelling1933analysis}: 
\begin{equation}
\label{eq:pcadefinition}
\pi_\text{PCA}(P) = {\cal{M}} P = p \;,
\end{equation}
where ${\cal{M}}$ is a $d \times D$ matrix   
consisting of the first $d$ eigenvectors (used as row vectors) that result from the diagonalization in PCA, i.e., in our typical use case of 2D visualization, these are the two eigenvectors corresponding to the two largest eigenvalues.


\begin{figure}[tb]
	\includegraphics[width = \linewidth]{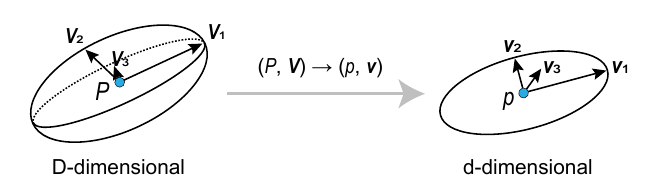}
	\vspace{-6mm}
	\caption{\label{fig:transformspaces} The local subspace of a point $P$ in the original space is spanned by basis vectors $\vec{V}_i$, and these are transformed to the $d$-dimensional visualization space, resulting in $p$ and $\vec{v}_i$, respectively.}
	\vspace{-6mm}
\end{figure}

\subsection{Local Subspaces and Transformation of Vectors}
Let a local subpace $S$ within the high-dimensional space be anchored around a point at position $P \in \RR^D$ and have dimensionality $L \leq D$. As shown in~\autoref{fig:transformspaces}, the subspace is spanned by the set of basis vectors ${\cal V} = \{\vec{V}_i\}_{i=1, \dots L}$:   
\begin{equation}
S = \left\{ Q \in \RR^D \; \bigg| \; Q = \sum_{i=1}^{L} \alpha_i \vec{V}_i + P \; \text{with} \; \alpha_i \in \RR \right\}   \;,
\end{equation}
\yh{where $L$ is the intrinsic dimensionality and the set of basis vectors ${\cal V}$  is obtained by computing local PCA~\cite{zhan2009adaptive}. Specifically, we first find the $k$ nearest neighbors of $P$, then  perform PCA with the point $P$ and its $k$ nearest neighbors, and finally construct ${\cal V}$ by picking the first $L$ eigenvectors that can explain most of total variance.}


To analyze the effect of $\pi$ on the local subspace, we reduce this problem to understanding how vectors are mapped from the high-dimensional space to the low-dimensional visualization space. Let us consider a vector $\vec{V} \in \RR^D$ anchored at point $P$. This vector is transformed \cite[page 290]{Goldstein1980mechanics} according to
\begin{equation}
\label{eq:vectortransformation}
\vec{v}  = \frac{\partial \pi(P)}{\partial P} \vec{V}  \;,
\end{equation}
with the Jacobian matrix of size $d \times D$,
\begin{equation}
\frac{\partial \pi(P)}{\partial P}   =
\begin{pmatrix}
\frac{\partial \pi(P)_1}{\partial P_1} & \dots  &  \frac{\partial \pi(P)_1}{\partial P_D}  \\
\vdots & \ddots & \vdots \\
\frac{\partial \pi(P)_d}{\partial P_1} &   \dots     & \frac{\partial \pi(P)_d}{\partial P_D}
\end{pmatrix} \;.
\end{equation}
The subscripts refer to the respective component of the Cartesian point.

It is important to point out that a vector $\vec{V}$ only ``lives'' together with an anchor point $P$, i.e., the two together form a fiber bundle \cite{Husemoller1994fibre}---here, a vector bundle or vector field.
In summary, the combined multidimensional projection of points and attached vectors can be written as:
\begin{equation}
\label{eq:basicvectortransformation}
\boxed{(P, \vec{V}) \mapsto (p, \vec{v})  = \left(\pi(P),  \frac{\partial \pi(P)}{\partial P} \vec{V}\right)}  \;.
\end{equation}
\autoref{fig:transformspaces} illustrates the geometry of points and vectors in both the original $D$-dimensional space and the $d$-dimensional visualization space.

\autoref{eq:basicvectortransformation} can be applied to all basis vectors $\vec{V}_i  \in {\cal V}$ of the local subspace. However, it should be noted that the transformed basis vector do not necessarily form a basis in the target space $\RR^d$. Often, the original subspace has dimensionality $L$ larger than the available dimensionality $d$ in the target space. Nevertheless, we can use the set of transformed vectors to learn about the characteristics of the multidimensional projection; see later in \autoref{sec:visnUI} and \autoref{sec:example}. This problem does not arise for $L \leq d$; here, the transformed basis vector(s) may indeed form a basis in the target space. A simple example is $L=1$, i.e., when just a single vector is transformed to show the effect on local linear correlation.

\subsection{Direct Computation of Vector Transformation}

An open question is the actual computation of the vector transformation from \autoref{eq:basicvectortransformation}. For a few cases, this computation is straightforward. For the example of PCA (\autoref{eq:pcadefinition}), we have a simple linear map $\pi$ and, therefore, the Jacobian matrix is identical to the matrix from the linear PCA map:
\begin{equation}
(P, \vec{V}) \mapsto (p, \vec{v})  = \left({\cal{M}} P , {\cal{M}} \vec{V}  \right)  \;.
\end{equation}
The above computation is valid for any linear projection.

However, multidimensional projections, in general, are nonlinear and more complex. And they may not lend themselves to an analytic derivations of the Jacobian matrix. Here, we could resort to using a numerical computation of the partial derivatives for the Jacobian matrix. The standard approach employs finite differences \cite{Press1992numrec}, for example, in the form of forward differences applied to each element of the Jacobian matrix:
\begin{equation}
\label{eq:forwarddifferences}
\frac{\partial \pi(P)_i}{\partial P_j} \approx \frac{\pi(P + h \, \vec{e}_j)_i -  \pi(P)_i}{h} \;,
\end{equation}
with the unit vector $\vec{e}_j$ in direction of the $j$-th dimension and a small distance measure $h \in \RR^+$. Forward differences provide a first-order approximation; second-order approximation is achieved by analogous central differences.
%
%
%

While finite differences are a valid means of approximating the vector transformation, they came with a number of shortcomings: They are just a numerical approximation and they need to compute the projection at a number of additional points $P + h \, \vec{e}_j$, which are not in the original set of data points. The latter issue has two negative implications. First, we need more computations of projections, which incurs higher computational costs. Second, and more importantly, DR techniques are data-dependent and often work only on the given input data; therefore, it can be hard to feed in data points that are different from the original point set or these additional points will modify the projection itself, which would lead to systematic errors.

To resolve these issues we now introduce a new vector transformation approach based on the implicit function theorem.

\subsection{Implicit Vector Transformation}

Nonlinear projection methods typically find the relationship between $P$ and $\pi(P)$ by optimizing an objective function $f(P, \pi(P))$. 
\rev{Here, the mapping between $P$ and $\pi(P)$ is implicitly described by $f$.}
However, even the simpler linear projections can be formulated in this way.
Therefore, we assume that any reasonable projection finds an optimal target location for each data point $P$ by solving the following optimization problem to arrive at the projected data points $\pi(P)$:
\begin{equation}
\label{eq:minimization}
\min_{\pi(P)} f(P, \pi(P)) \; .
\end{equation}
In other words, any resulting point $\pi(P)$ ``sits'' in a local minimum with respect to the cost function. If this was not the case, the projection method could and should move the projected point $\pi(P)$ to reduce the cost. 

\rev{\autoref{eq:minimization} leads to the necessary condition for the minimum if $f$ is a smooth function, and the minimum is inside (not on boundaries of the domain):}
\begin{align}
\centering
& \frac{\partial}{\partial{\pi(P)}}f(P, \pi(P))= 0\;, \quad\text{where}\nonumber\\
&\frac{\partial}{\partial \pi(P)} = \left[\frac{\partial }{\partial \pi(P)_1},\frac{\partial }{\partial \pi(P)_2},\cdots, \frac{\partial}{\partial \pi(P)_d}\right]^T\;.
\label{eqn:nonlinObj}
\end{align}
Using the implicit-function differentiation theorem~\cite[Chapter~11]{Loomis1990calculus},
we take the partial derivative with respect to $P$ on both sides of Equation~\ref{eqn:nonlinObj}, and apply the chain rule:
\begin{align}
&\frac{\partial^2 f(P,\pi(P))}{\partial P\partial \pi(P)}  + \frac{\partial^2 f(P,\pi(P))}{\partial \pi(P)^2} \frac{\partial \pi(P)}{\partial P} = 0\\
\Rightarrow\quad&\boxed{\frac{\partial \pi(P) }{\partial P} = -\left(\frac{\partial^2 f(P, \pi(P))}{\partial \pi(P)^2} \right)^{-1}\left[\frac{\partial^2  f(P, \pi(P))}{\partial P\partial \pi(P)}\right]}\;.
\label{eqn:nonlinDiff}
\end{align}
\autoref{eqn:nonlinDiff} is the key mathematical result of our paper. It describes the implicit vector transformation that we can now use to transform subspaces. The transformation is generally applicable and completely accurate as long as the reasonable assumption from \autoref{eq:minimization} holds for the underlying smooth multidimensional projection. Furthermore, the transformation has to be computed only for the original points $P_i$ in the dataset, and not at any other locations or for any other points.

For a specific choice of nonlinear dimensionality projection method, we have to evaluate the two derivatives $\left(\frac{\partial^2 f(P, \pi(P))}{\partial \pi(P)^2}\right)^{-1}$ and $\frac{\partial^2  f(P, \pi(P))}{\partial P\partial \pi(P)}$ with the specific $f(P,\pi(P))$ used in the method.

\subsection{Application to MDS}
Here, we briefly explain the objective functions and the strategy to compute~\autoref{eqn:nonlinDiff} for the representative nonlinear method MDS.  We have also derived solutions for t-SNE that can be found in the supplemental material.

The MDS method has both linear and nonlinear versions. Here, we consider the SMACOF version~\cite{borg2005modern}, which is widely used as the default method in data analysis packages, for example, scikit-learn, due to its good performance.
SMACOF minimizes the objective function:
\begin{small}
\begin{equation}
F = \frac{1}{2}\sum_{i=1}^n\sum_{j=1}^{n}(||x_i-x_j||-||y_i-y_j||)^2\;,
\label{eqn:SMACOF}
\end{equation}
\end{small}
where $x_i$, $x_j$ are original high-dimensional data points, and $y_i$ and $y_j$ are projected low-dimensional data points. 
We can map this equation to our original formulation in \autoref{eqn:nonlinDiff} by associating $F$ with $f$, $x_i, x_j$ with $P$, and $y_i, y_j$ with $\pi(P)$. We first compute the partial derivative $\frac{\partial F}{\partial y_i}$ of~\autoref{eqn:SMACOF}:
\begin{small}
\begin{equation}
\frac{\partial F}{\partial y_{i}}=2\sum_{j \neq i}\left(1-\frac{\left\|x_{i}-x_{j}\right\|}{\left\|y_{i}-y_{j}\right\|}\right)\left(y_{i}-y_{j}\right),
\label{eqn:mdspartial}
\end{equation}
\end{small}
where the denominator requires all duplicate points to be removed.

From $\frac{\partial F}{\partial y_{i}}$, we then derive the second derivatives $\frac{\partial^2 f(P, \pi(P))}{\partial \pi(P)^2} = \frac{\partial^2 F}{\partial y_i \partial y_j}$ and $\frac{\partial^2  f(P, \pi(P))}{\partial P\partial \pi(P)} = \frac{\partial^2 F}{\partial y_i \partial x_j}$ for cases $i = j$ and $i \neq j$, respectively:
\begin{small}
\begin{equation}
\frac{\partial^{2} F}{\partial y_{i} \partial y_{j}}=
\begin{cases}
2 \sum_{k \neq i}\left(\left(1-\frac{\left\|x_{i}-x_{k}\right\|}{\left\|y_{i}-y_{k}\right\|}\right) I+\frac{\left\|x_{i}-x_{k}\right\|}{\left\|y_{i}-y_{k}\right\|^{3}}\left(y_{i}-y_{k}\right)\left(y_{i}-y_{k}\right)^{T}\right)
,&\text { if\quad} i=j \\
2\left(\frac{\left\|x_{i}-x_{j}\right\|}{\left\|y_{i}-y_{j}\right\|}-1\right) I-\frac{2\left\|x_{i}-x_{j}\right\|}{\left\|y_{i}-y_{j}\right\|^{3}}\left(y_{i}-y_{j}\right)\left(y_{i}-y_{j}\right)^{T}
,&\text { otherwise\;, }
\end{cases}
\end{equation}
\end{small}
where $I$ is the identity matrix, and
\begin{small}
\begin{equation}
\frac{\partial^{2} F}{\partial y_{i} \partial x_{j}}=
\begin{cases}
\sum_{k \neq i}-2 \frac{\left(y_{i}-y_{k}\right)}{\left\|y_{i}-y_{k}\right\|}  \frac{\left(x_{i}-x_{k}\right)^{T}}{\left\|x_{i}-x_{k}\right\|}\;, &\text{if\quad} i=j \\
2 \frac{\left(y_{i}-y_{j}\right)}{\left\|y_{i}-y_{j}\right\|} \frac{\left(x_{i}-x_{j}\right)^{T}}{\left\|x_{i}-x_{j}\right\|}
\;,& \text{otherwise\;.}
\end{cases}
\end{equation}
\end{small}

\section{Visualization and Interaction}
\label{sec:visnUI}
In this section, we elaborate on how to visualize projected local subspaces with glyphs, explain specialized user interactions for visual exploration, and briefly report on the implementation of our method.
\autoref{fig:ui} shows a screenshot of our interactive tool.

\subsection{Glyph Generation}
\label{sec:glyph}

We use a glyph to visualize the transformation from the $L$-dimensional local linear subspace embedded in the original $D$-dimensional space to $d$-dimensional visualization space.
The original basis vectors $\vec{V}_i$ are the eigenvectors that come from local PCA in high-dimensional space, and are assumed to be normalized to unit length. We incorporate a weight $\alpha_i$ that measures the importance of the eigenvalue $\lambda_i$ associated with $\vec{V}_i$, according to $\alpha_i = \lambda_i / \sum_{j=1}^{L}\lambda_j$. Therefore, our original basis is $\{\alpha_{i}\vec{V}_i | i = 1, \dots, L\}$, anchored at point $P$. According to \autoref{eq:basicvectortransformation}, it is transformed to $(\pi(P), \{\alpha_{i}\vec{v}_i |i = 1, \dots, L\})$. The latter one is what the glyph should visualize.

\begin{figure}[htb]
	\centering
	\includegraphics[width=0.95\linewidth]{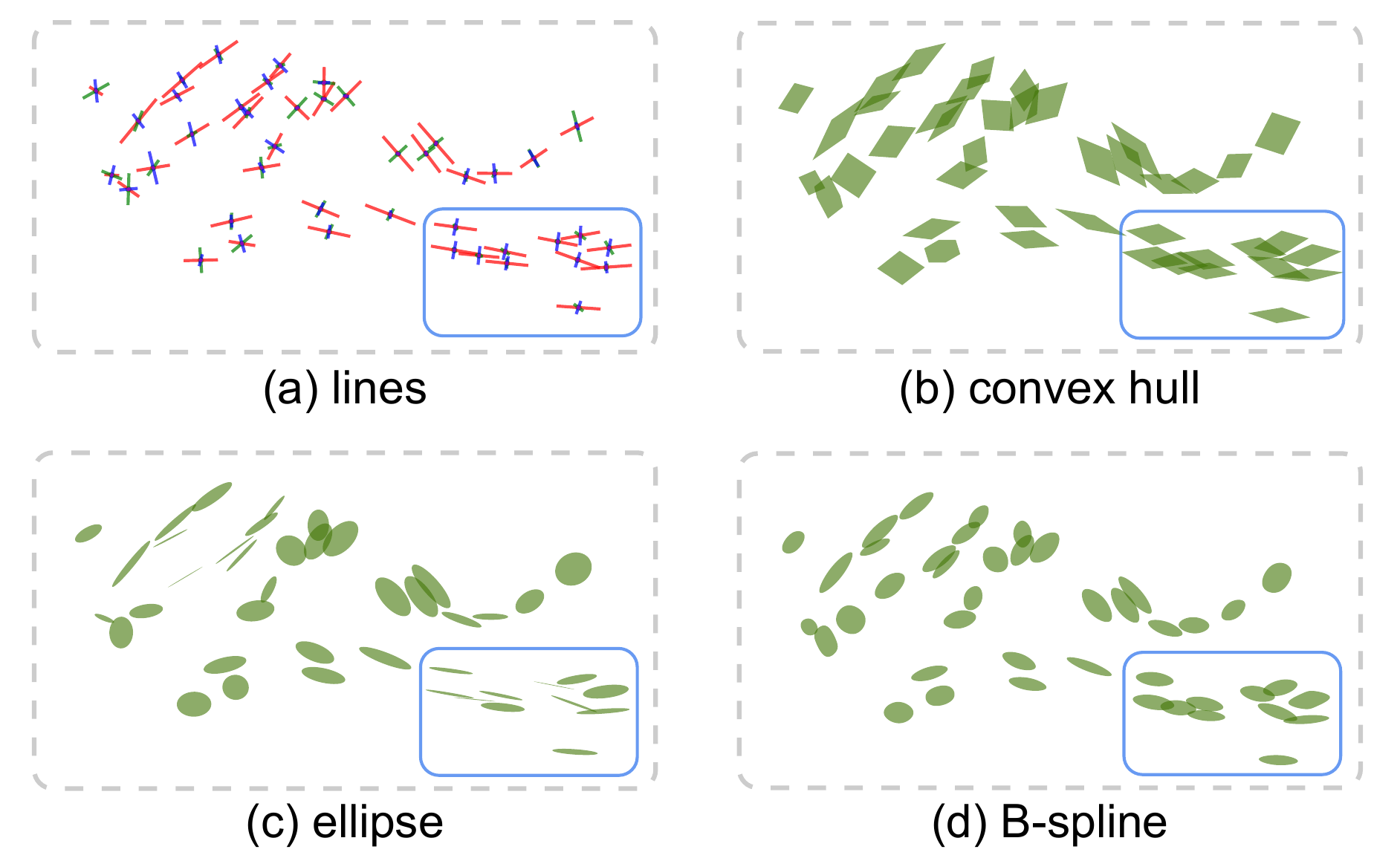}
	\vspace{-3mm}
	\caption{\rev{Different options of glyph visualizations encoding transformed basis vectors: (a) lines, (b) convex hull, (c) ellipse, and (d) B-spline.}}
	\label{fig:glyphDesign}
	\vspace{-4mm}
\end{figure}

\begin{figure}[htb]
	\centering
	\includegraphics[trim={1cm 1cm 1cm 1cm}, clip, width=\linewidth]{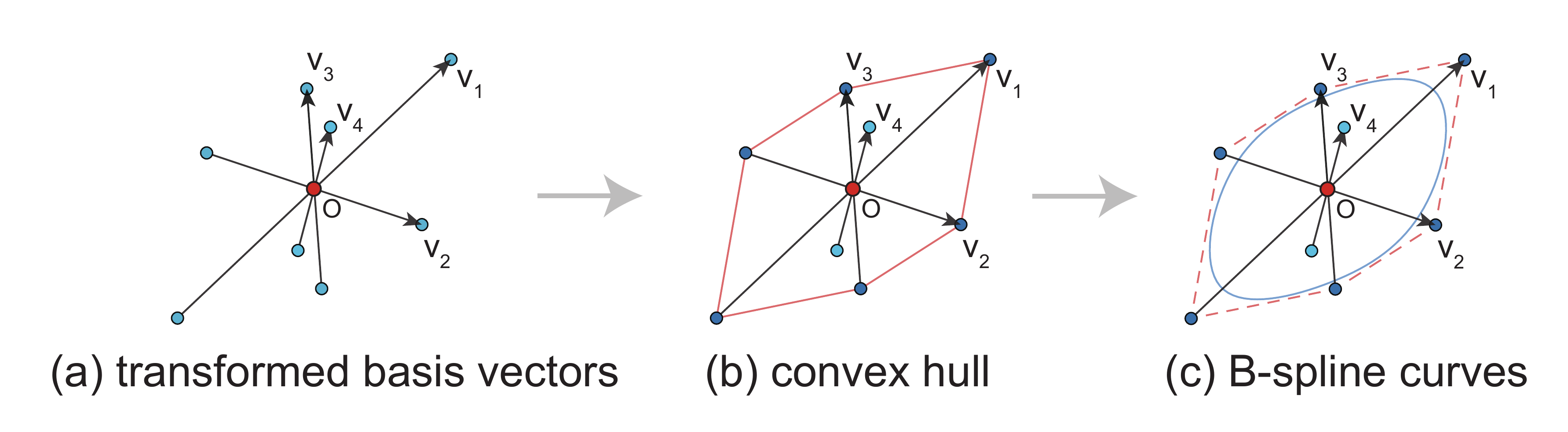}
	\vspace{-6mm}
	\caption{A glyph is generated for (a) the set of transformed basis vectors: the glyph (c) is computed using B-spline of (b) the convex hull of the transformed basis vectors.}
	\label{fig:glyphGen}
	\vspace{-2mm}
\end{figure}

However, naive visualization is not possible because the $\alpha_{i}\vec{v}_i$ usually do not form a basis in the $d$-dimensional visualization space. Typically $L > d$ and, thus \yh{direct visualization of all $L$ vectors as lines (\autoref{fig:glyphDesign}~(a)) might cause visual clutter that impairs perception. Ellipse based encoding can result in smooth visualizations (\autoref{fig:glyphDesign}~(c)), but it supports only two transformed basis vectors. Although the convex hull (\autoref{fig:glyphDesign}~(b)) of the transformed basis vectors can show all $L$ basis vectors, it is not smooth and overlapping areas of neighboring glyphs are large and potentially misleading.
Given the disadvantages of these options, we design a closed B-Spline convex hull glyph, as shown in \autoref{fig:glyphGen}~(d), which can represent the major transformed basis vectors and it is smooth and easy to distinguish.}

%
%
%

Our solution to this problem is illustrated in~\autoref{fig:glyphGen}:
a glyph is generated by first computing the convex hull (\autoref{fig:glyphGen}~(b)) of the transformed vectors centering at the projected point $\pi(P)$ (\autoref{fig:glyphGen}~(a)); then, a smooth shape is computed using a B-spline~\cite{de1978practical} within the convex hull whose vertices are used as control points (\autoref{fig:glyphGen}~(c)).

The B-spline is used because of its desired properties: it stays within the convex hull and follows a local control of its shape, which ensures that glyphs are comparable.
One principal issue in glyph-based visualization is occlusion caused by the larger coverage on screen than by small dots.
We \yh{alleviate} the issue by using transparency and alpha blending: glyphs are assigned user-specified opacities, and they  are drawn in layers such that larger glyphs are drawn below smaller ones.
\begin{figure*}[htb]
	\centering
	\includegraphics[width=0.8\linewidth]{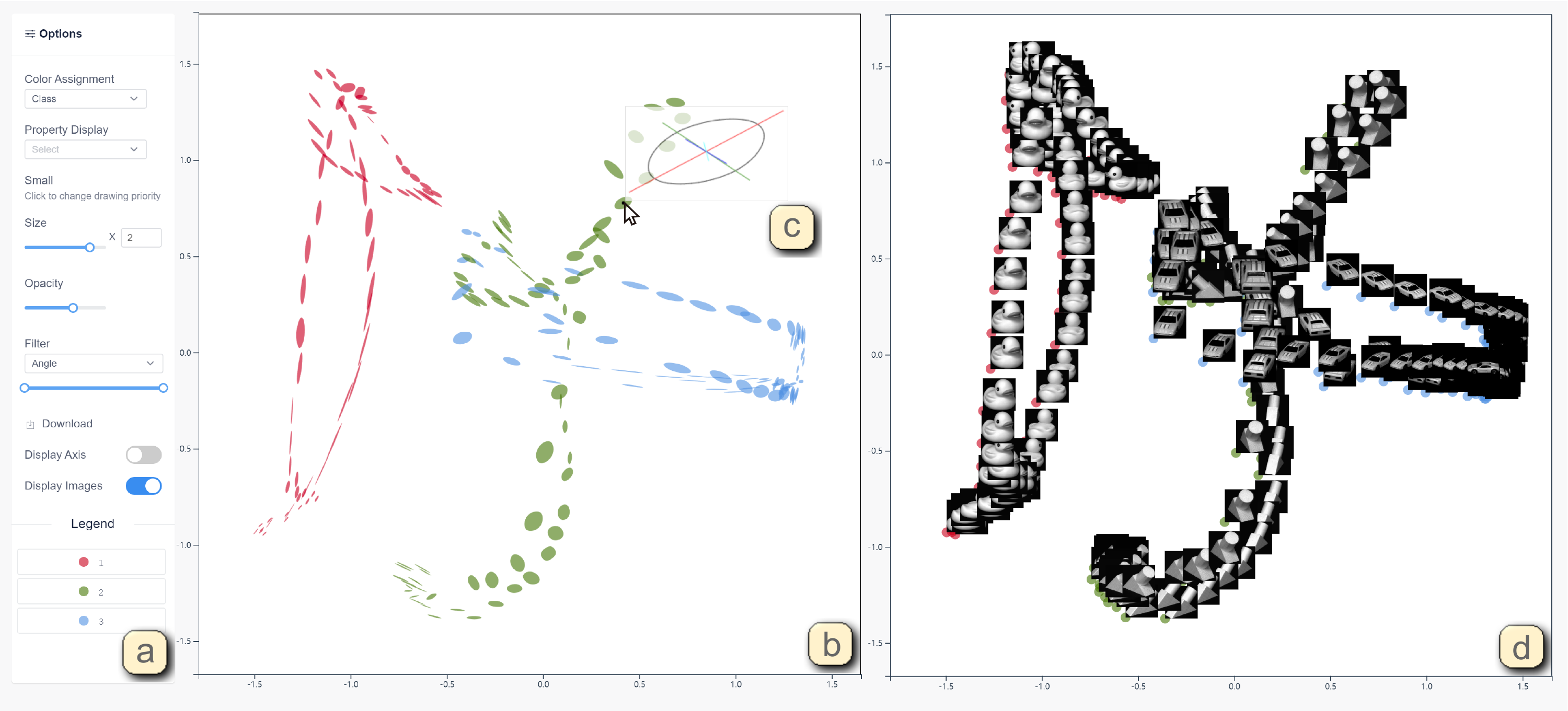}
	\vspace{-3mm}
	\caption{Main modules in our web-based interactive visualization system: (a) control panel, (b) local subspace projection view with (c) zoomable glyphs of interest, and (d) point-based dimensionality reduction view with option to show image data for reference. This example shows the COIL dataset with PCA.}
	\label{fig:ui}
	\vspace{-5mm}
\end{figure*}



Furthermore, scaling is supported in our visualization tool so that the user can change the size of glyphs globally.
The visualization is generated by alpha-blending glyphs assigned with desired visual properties---color, opacity, and scale.
These visual properties can be interactively modified by the user with interactions as shown on the left-hand side of~\autoref{fig:ui}.

\subsection{Interaction}
Our interactive exploration system is shown in~\autoref{fig:ui}. It comprises three modules: the control panel (\autoref{fig:ui}~(a)), the local subspace projection viewer (\autoref{fig:ui}~(b)) with glyph zoom-in (\autoref{fig:ui}~(c)), and a linked point-based projection viewer with option to show image references (\autoref{fig:ui}~(d)) for comparison.
The local subspace projection viewer is synchronized with the point-based viewer for zooming-and-panning---it is particularly useful for comparing local trends of glyphs and changes in image data (projection viewer) in high-dimensional datasets in computer vision as shown in Section~\ref{sec:example}.
Brushing-and-linking is supported for closer examinations: the user can select glyphs (in~\autoref{fig:ui}~(b)) or points (in~\autoref{fig:ui}~(d)) of interest with a lasso and corresponding points and glyphs are highlighted.

Furthermore, a set of specialized user interactions---including projection-quality filtering, flexible color mapping, and glyph structure zoom-in---are adopted in our method.
The user changes opacity and size of glyphs with two sliders to reduce the occlusion for further explorations of local subspaces.
Color mapping is useful when visualizing various metrics of data points and their local subspaces---for example, class, and the projection quality (see Section~\ref{sec:projError}).
The user may be also interested in the actual projected vectors from the local subspace.
This information is especially useful for identifying anomalies.
A zoom-in visualization of the glyph (\autoref{fig:ui}~(c)) shows transformed basis vectors (red, green, blue, and cyan are for the first through fourth basis vector, respectively) inside the outline of the B-spline shape.
The zoom-in is activated whenever the mouse pointer hovers over a glyph.
Glyph filtering allows the user to focus on glyphs based on a user-selected metric among desired class, anisotropy, or projection quality.

\rev{These interactions allow us to adopt the ``overview first, zoom and filter, then details-on-demand"~\cite{shneiderman1996eyes} process for visual analysis. 
	Specifically, we first observe the visualization of the whole dataset to find interesting general trends and local patterns: e.g., anomaly and crossings.
	Then, we focus on these patterns and examine them with interactive filtering and brushing-and-linking, and verify our findings with the original data if possible (for example, images); when necessary, we zoom in to further explore the details.}



\begin{figure}[htb]
	\centering
	\vspace{-6mm}
	\subfloat[projection error]{\includegraphics[width=0.47\linewidth]{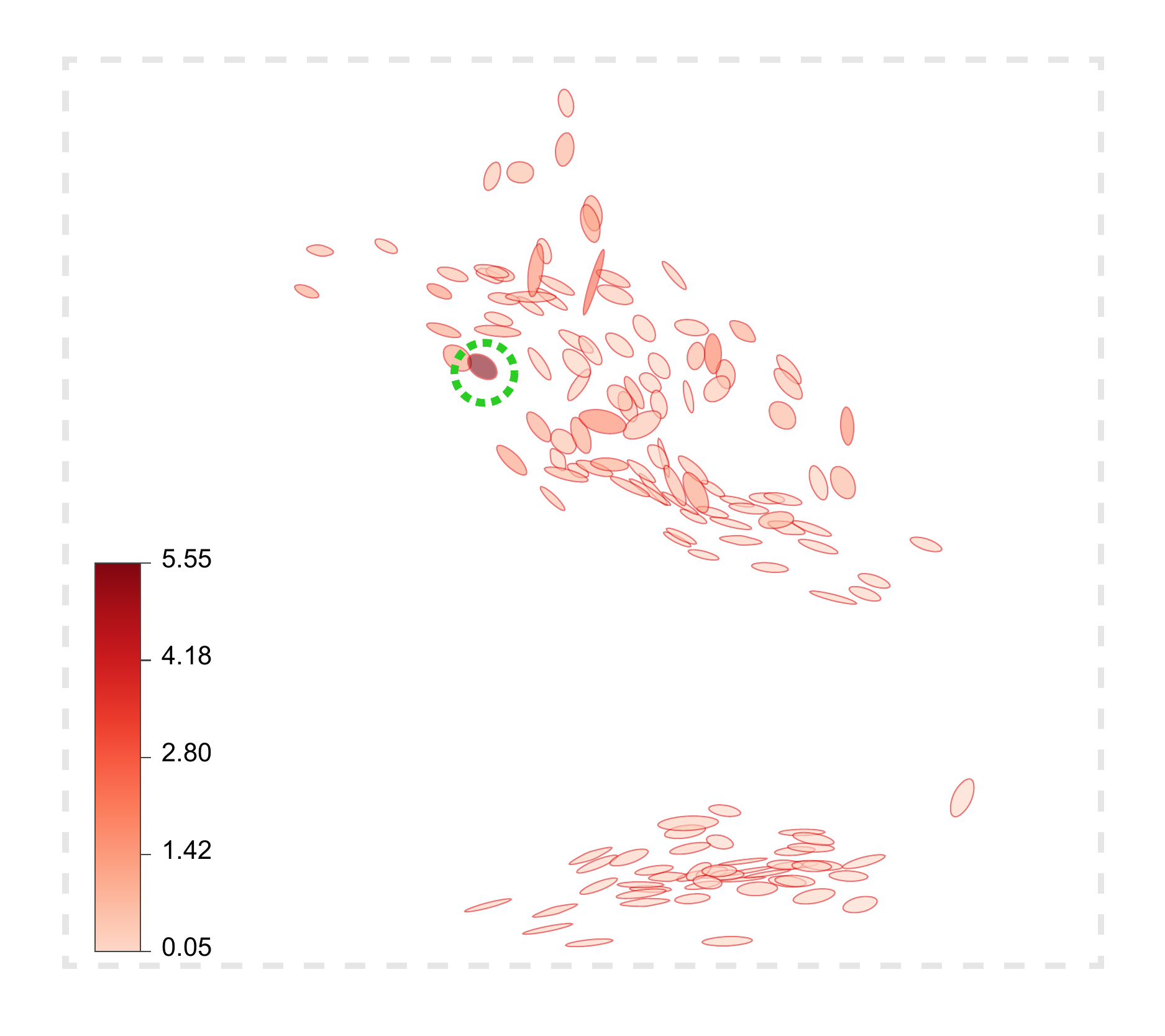}}\hfill
	\subfloat[neighborhood preservation]{\includegraphics[width=0.47\linewidth]{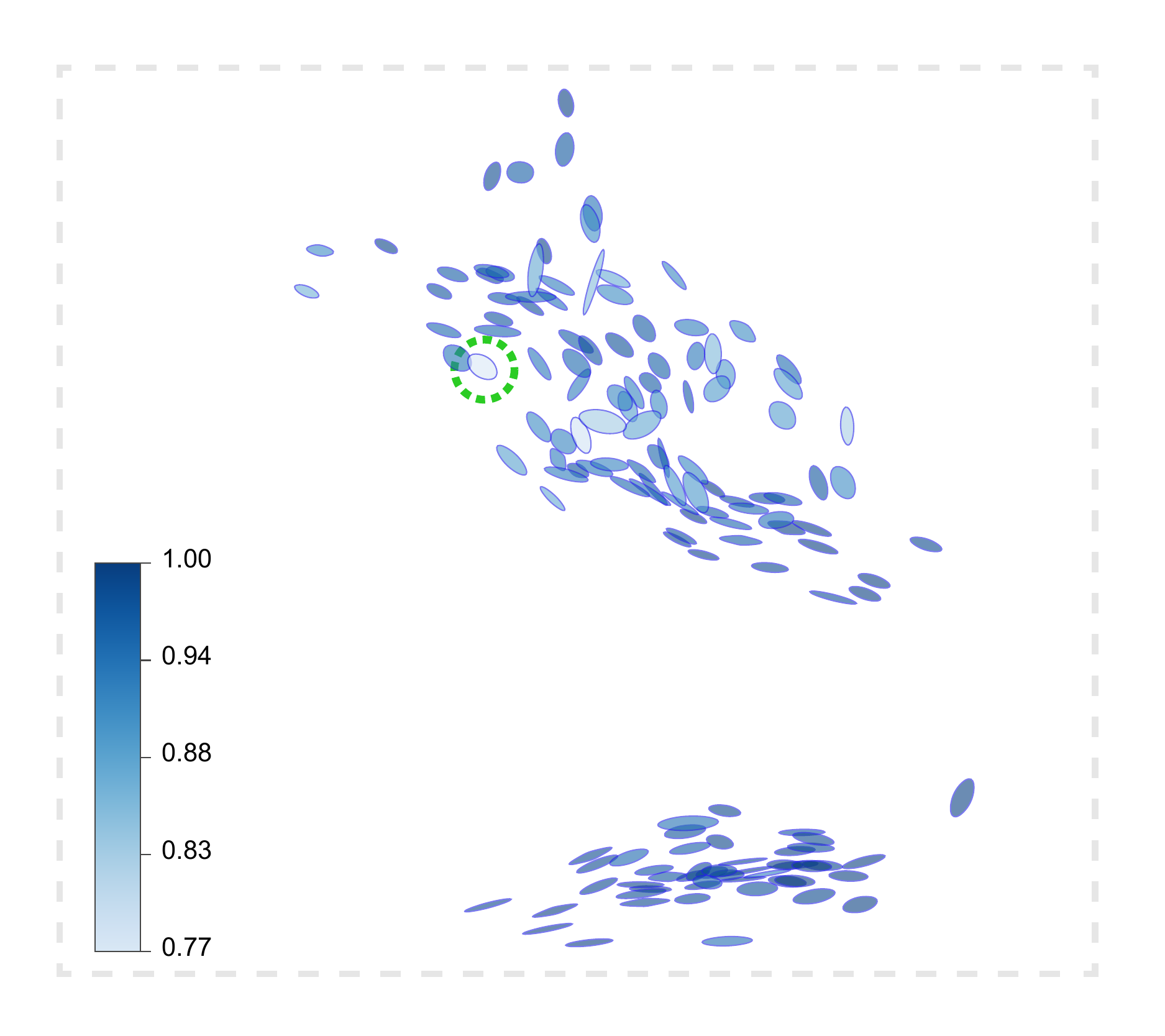}}
	\vspace{-2mm}
	\caption{\rev{Color mapping of two projection quality metrics for MDS-projected Iris dataset: (a) stress error where a lower score indicates better projection quality, and (b) trustworthiness where a higher score indicates better projection quality. The point highlighted by the green circle has the largest stress error and least trustworthiness.}}
	\label{fig:linearity}
	\vspace{-4mm}
\end{figure}

\subsubsection{Projection Quality}
\label{sec:projError}
The quality of the DR method used for data points is also important for visual analysis.
We introduce three metrics to describe the projection quality:  projection error, neighbor preservation, and linearity.

The projection error metric is the loss function of the particular DR method (\autoref{fig:linearity}~(a)). For MDS, the metric is the stress error.
The neighborhood preservation degree is independent of the DR method, \yh{measured by
the trustworthiness~\cite{Espadoto19,Nonato19} (\autoref{fig:linearity}~(b)), where $k$ is set to be the same as the one used for computing local PCA.}

The linearity metric describes the anisotropy of the local subspace in the original dimensionality---\rev{the ratio of the magnitude between largest eigenvalue and second-largest eigenvalue.} Unlike the above two metrics, coloring the linearity with each glyph only indicates how similar the glyph-encoded local subspaces in the original high-dimensional space are, but the change-of-linearity can be revealed by looking at the aspect ratio of the glpyh itself. For example, the glyphs selected by the blue box in~\autoref{fig:compare} has linearity close to one, but its result shape is elongated in~\autoref{fig:compare}~(b).

The conjunction of these metrics help us better understand behaviors of the DR method.
For example, it seems that the projection error \autoref{fig:linearity}~(c)) is negatively correlated with the neighborhood correlation (\autoref{fig:linearity}~(d)), especially for the highlights with orange circles, in the MDS projection.

\subsection{Implementation}
The computational steps in our method---vector transformation by implicit differentiation and multidimensional projections---were implemented using Python and NumPy.
These steps are calculated only once and the results are used as input to our web-based user interface.
The user interface was implemented using JavaScript and Vue.js. The local subspace projection viewer is based on WebGL so that interactive visualization is achieved even for a large number of glyphs; the point-based projection viewer was realized using D3.
Thanks to our efficient visualization tool, full interactivity is achieved for all datasets used in our paper.

\section{Numerical Evaluation}
\label{sec:eval}
\begin{table}[htb]
	\centering
	\caption{Statistics of transformed basis vectors of non-border data points of the planar data.}
	\footnotesize
	\begin{tabular}{ccccc}
		&
		mean len1 &
		mean len2 &
		mean angle&
		std of angles\\
		\toprule
		Ours & 0.10068096 & 0.10068095 & 89.92020374 & 0.00502165\\
		Rand~\cite{faust2019dimreader}	& 0.10005878 & 0.09957406 & 89.03798816 & 0.50298048\\
		\bottomrule
	\end{tabular}\vspace{-1em}
	\label{tbl:verf}
\end{table}


We numerically compare our accurate implicit vector transformation method to the approximated random approach used in DimReader~\cite{faust2019dimreader}.
In contrast to our analytical technique, DimReader randomly chooses half the points and calculates one projected vector of each data point using auto-differentiation, and the process is repeated for the number of $L$ basis vectors.
\rev{We use the aforementioned implementation for our method and our own implementation of the random approach with publicly-available code snippets of DimReader on the Internet (both in Python without acceleration) for all evaluations.}	

\begin{figure}[htb]
	\centering
	\vspace{-1mm}
	\includegraphics[width=0.9\linewidth]{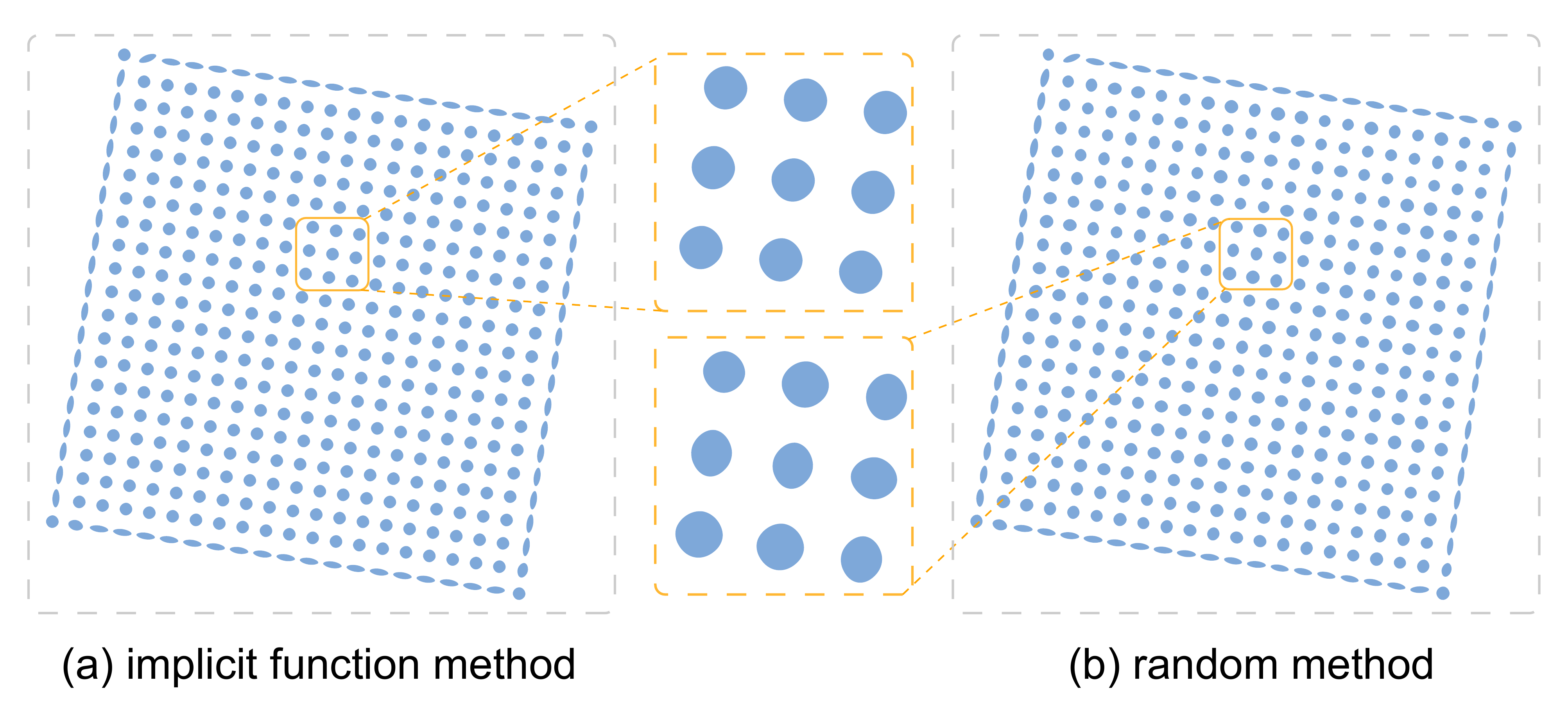}
	\caption{Transformed basis vectors visualized as glyph by (a) our implicit function method and (b) the random method~\cite{faust2019dimreader}. The data points are on a plane in 3D sampled using a regular grid, and projected to 2D with MDS.   }
	\label{fig:verf}
	\vspace{-2mm}
\end{figure}

\rev{
Note that our implicit function method is the analytical, i.e., accurate, transformation of basis vectors, whereas the random approach is an approximation.
To verify the correctness of our method, a synthetic planar data sampled on a 3D regular grid is generated and projected to 2D using MDS. 
With a neighborhood of $k = 8$, the transformed basis vectors of all non-border data points should be, theoretically, exactly orthogonal, i.e., 90 degrees, and having the same magnitudes. 
It can be seen in~\autoref{fig:verf} that glyphs of transformed basis vectors with our method appear identical (\autoref{fig:verf}~(a)), whereas different shapes of glyph are visible with the random approach (\autoref{fig:verf}~(b)). 
Quantitatively, Table~\ref{tbl:verf} summarizes basic statistics of transformed basis vectors of non-border data points of both methods: our method has a smaller difference of mean lengths of basis vectors (mean length 1 vs 2), a mean angle closer to 90 degrees (mean angle) with lower standard deviation (std of angles) than the random method. 
Furthermore, distributions of these measures are shown as histograms in the supplemental material. This verifies that our method is more accurate than the random method. 
 }

\begin{figure}[htb]
	\vspace{-2mm}
	\mbox{}\\[3ex]
	{\includegraphics[width = 0.98\linewidth]{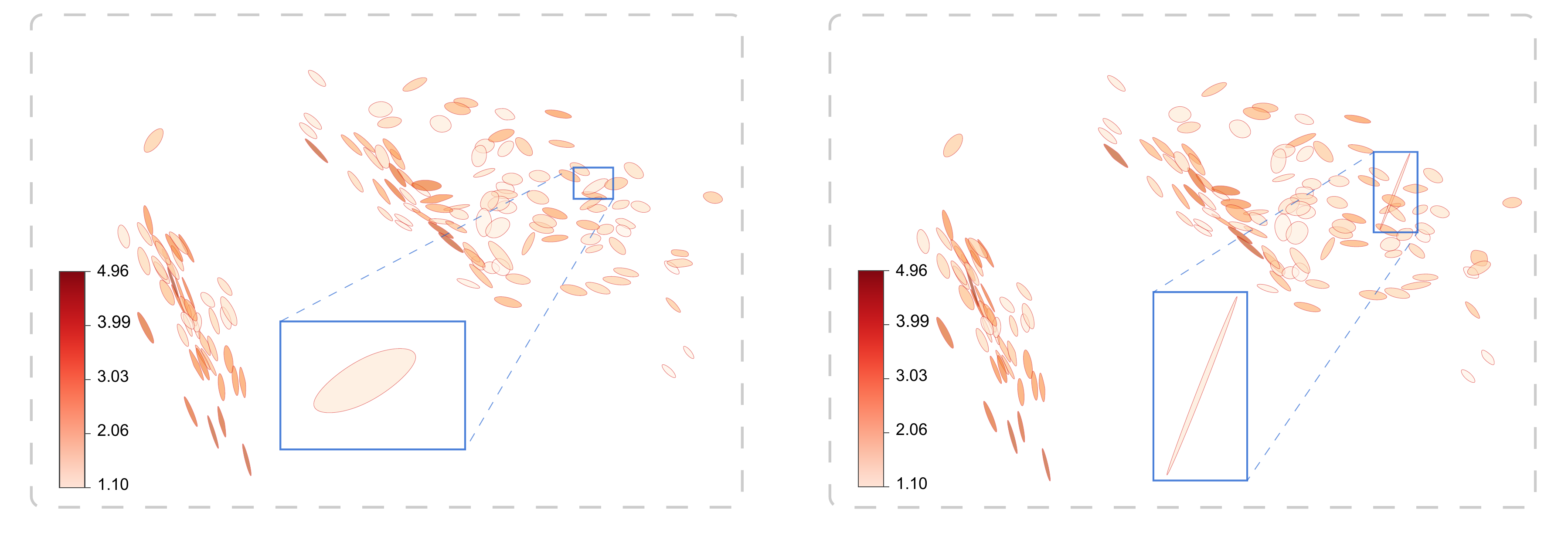}}\\[-2ex]
	\mbox{}\hfill\subfloat[implicit function method]{~~~~~~~~~~~~~~~~~~~~~~~~~~~~~~~~~~~~~~~~} \hfill\hfill \subfloat[random method]{~~~~~~~~~~~~~~~~~~~~~~~~~~~~~~~~~~~~~~~~} \hfill\mbox{}
	\caption{Glyphs with basis vectors transformed by our implicit function method (a) and the random method~\cite{faust2019dimreader} (b) for
			the MDS-projected Iris dataset. The color indicates the linearity of each glyph, while the shape indicates the change of the linearity in the 2D projection. 
			We highlight the long thin glyph (selected by the box) that resulted from the random method.  }
	\label{fig:compare}
	\vspace{-5mm}
\end{figure}

\rev{Next, we perform evaluations on real-world multidimensional datasets using nonlinear projections via MDS and t-SNE.}
We employ typical benchmark datasets: Iris (147 4D points---three duplicated points are removed), Wine (178 13D points), and Digits40 (606 40D points).
%
\autoref{fig:compare} compares the glyph visualizations of transformed basis vectors generated by our method and the random method on the Iris dataset color-coded with the linearity measure.
It can be seen that the random method generates more abnormal glyphs---with excessively long and thin shapes or with distinct directions inside a neighborhood of uniform directions.
\rev{We also measure computation times for the two methods for transforming the top five (four for the Iris data) basis vectors: the results show that our method is faster than the random method for all test datasets with both MDS and t-SNE projections (Table 2 of the supplemental material).
All figures of glyph visualizations generated during the evaluation can be found in the supplemental material.}
This is evidence that our implicit differentiation method is a fast and accurate vector transformation method from high-dimensional to low-dimensional space.

\newcommand{\multidWidth}{0.46\linewidth}
\begin{figure*}[htb]
	\centering
	\begin{tabular}{ccc}
		&
		Wine &
		Seeds\\[-2ex]
		\rot{90}{~~~PCA}&
		\subfloat{\includegraphics[width=\multidWidth]{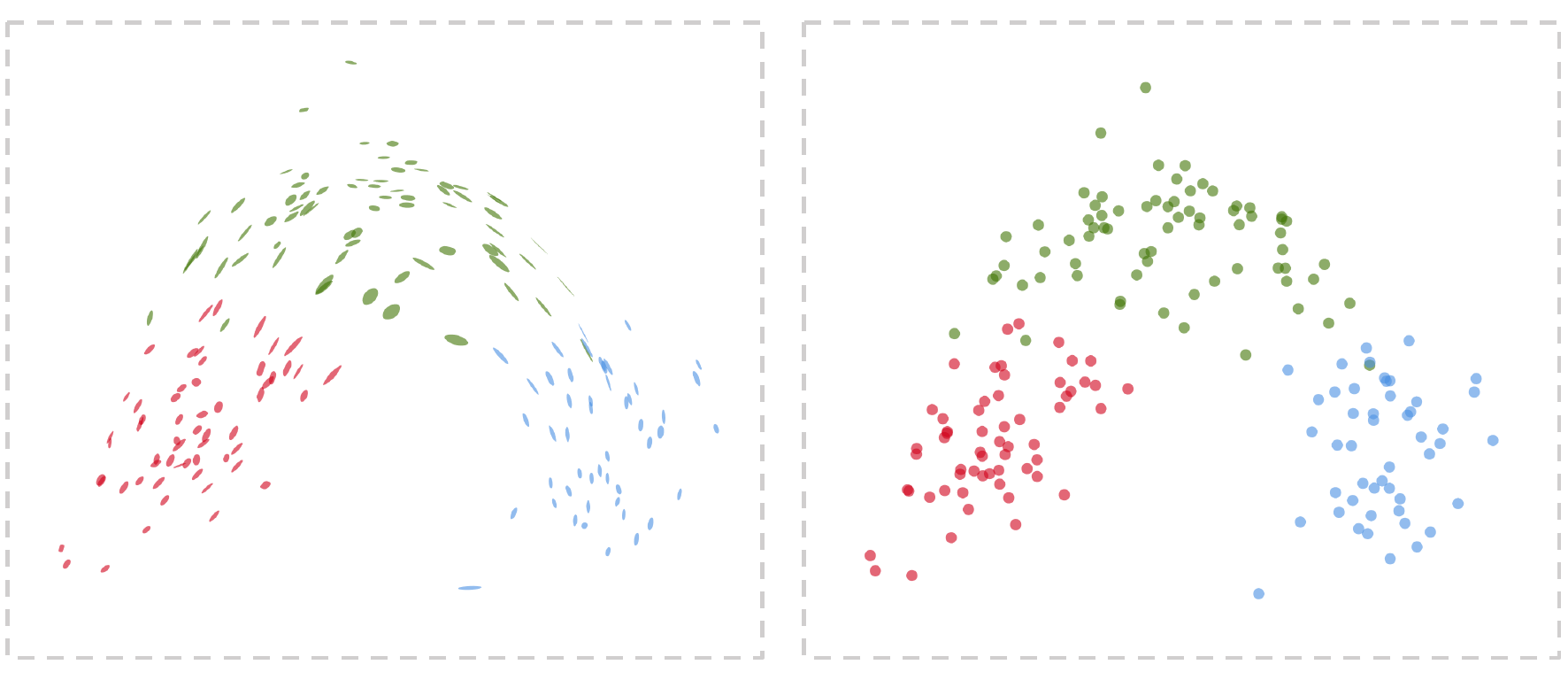}}&
		\subfloat{\includegraphics[width=\multidWidth]{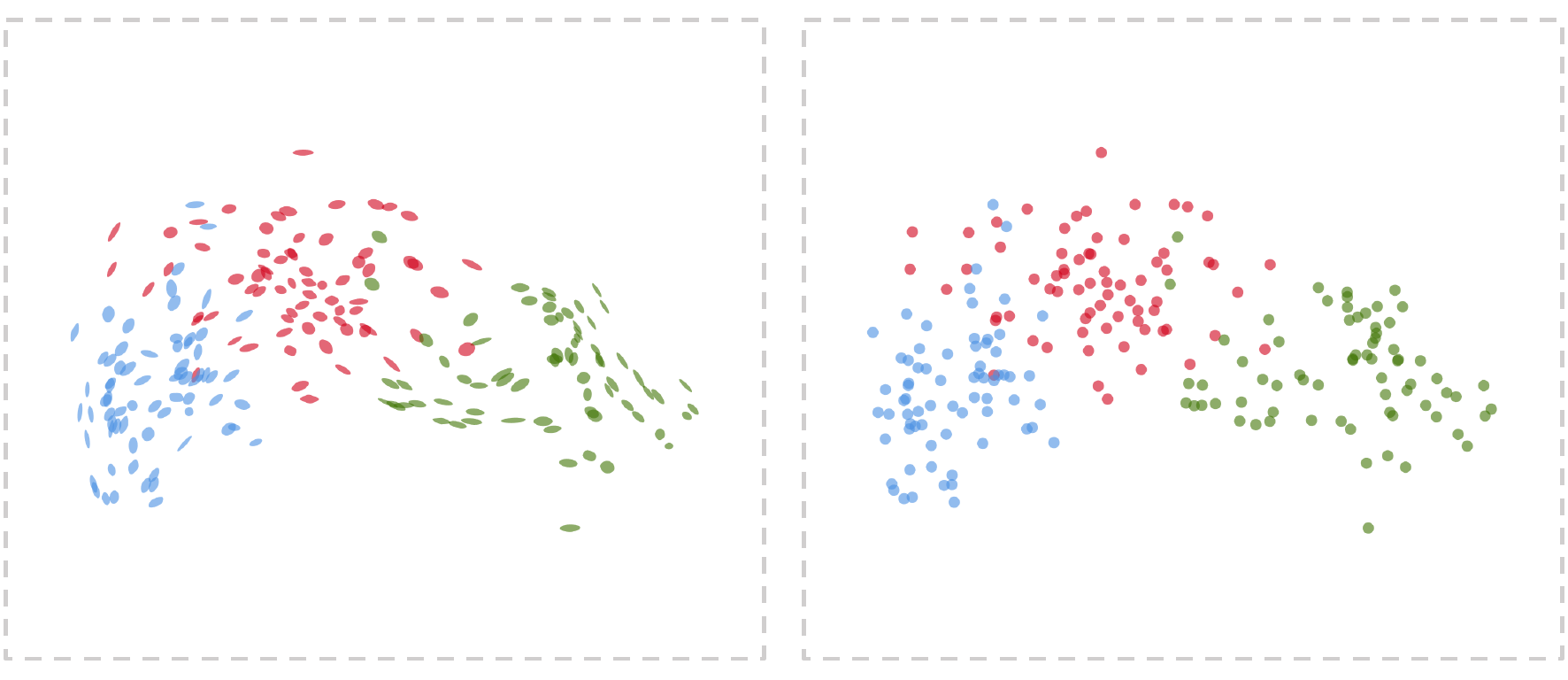}}\\	[-2ex]
		\rot{90}{~~~MDS}&
		\subfloat{\includegraphics[width=\multidWidth]{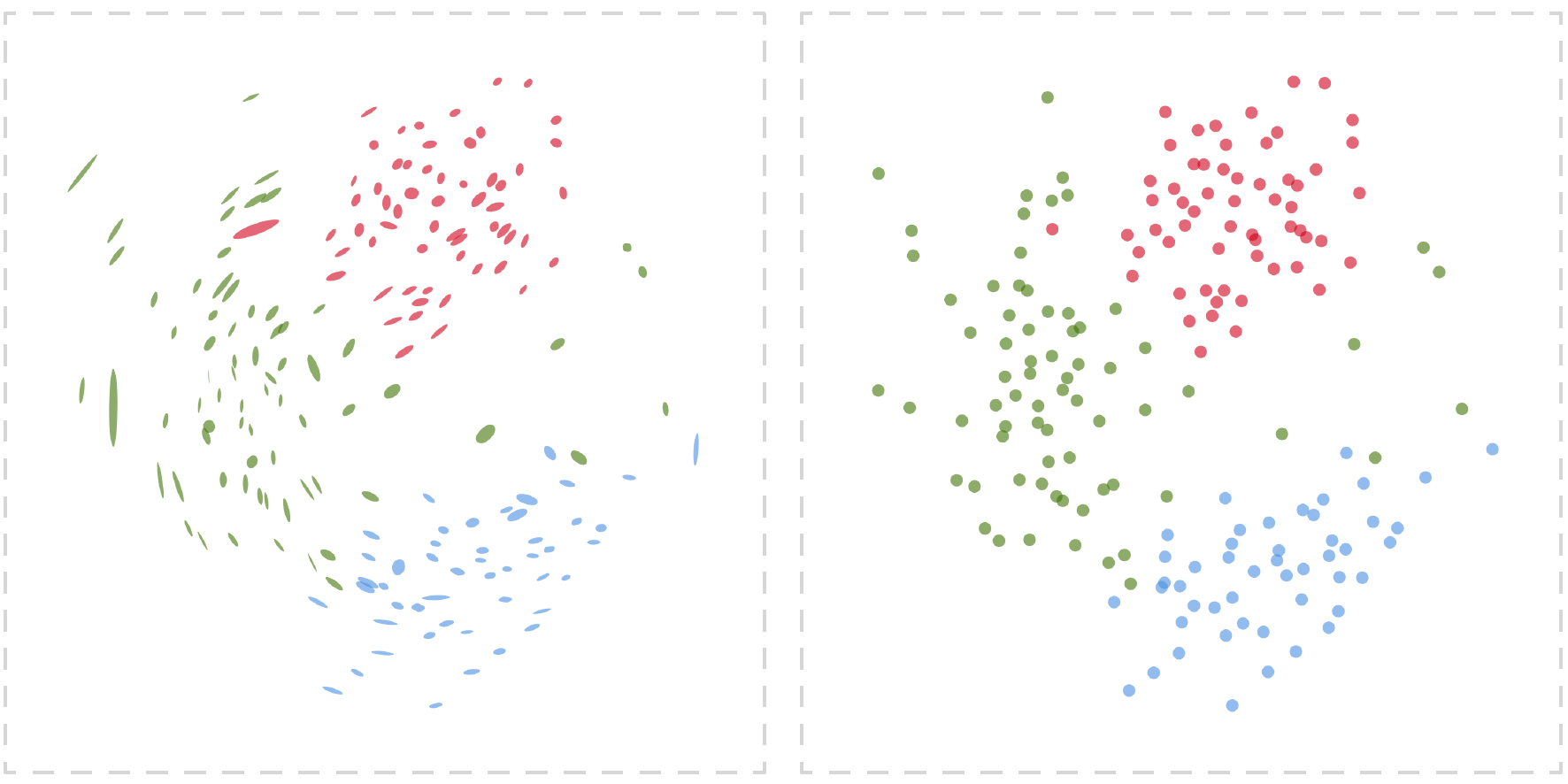}}&
		\subfloat{\includegraphics[width=\multidWidth]{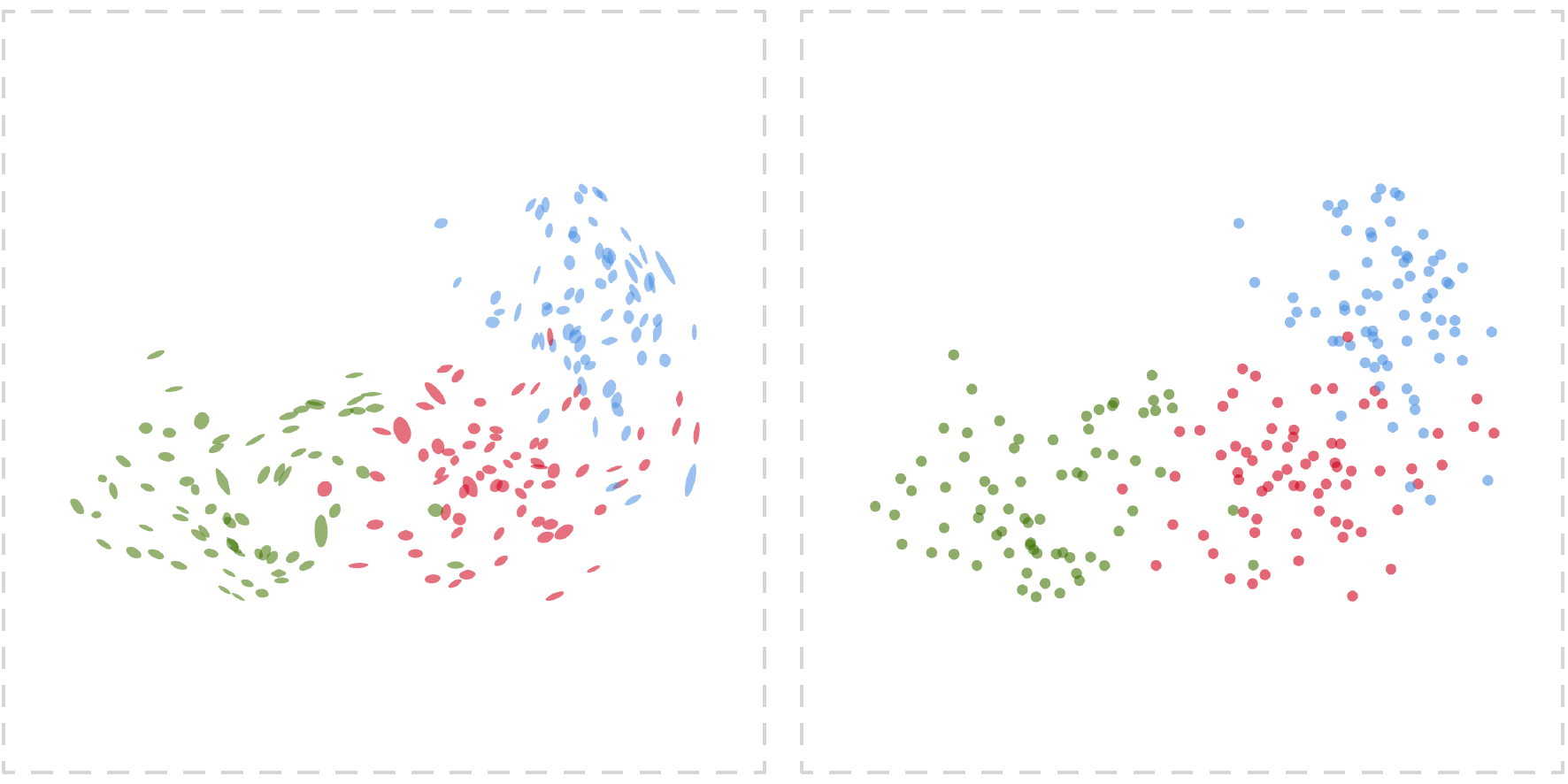}}	\\	[-2ex]
		\rot{90}{~~~t-SNE}&
		\subfloat{\includegraphics[width=\multidWidth]{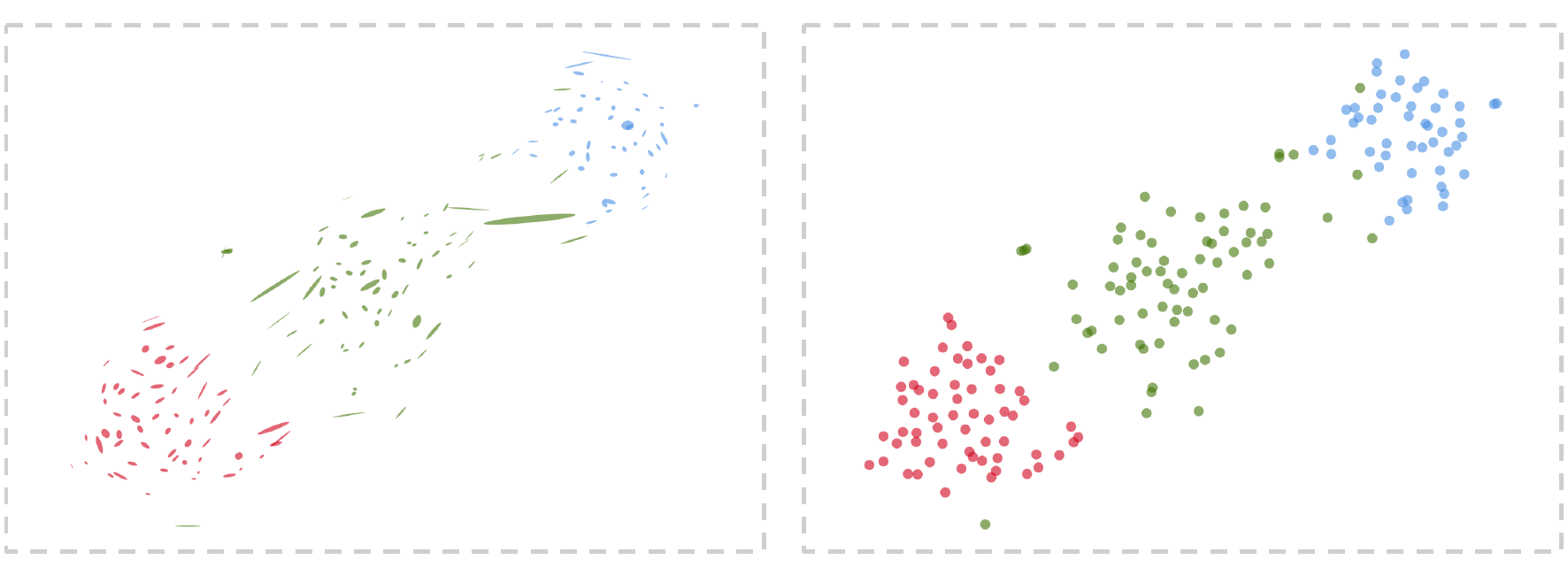}}&
		\subfloat{\includegraphics[width=\multidWidth]{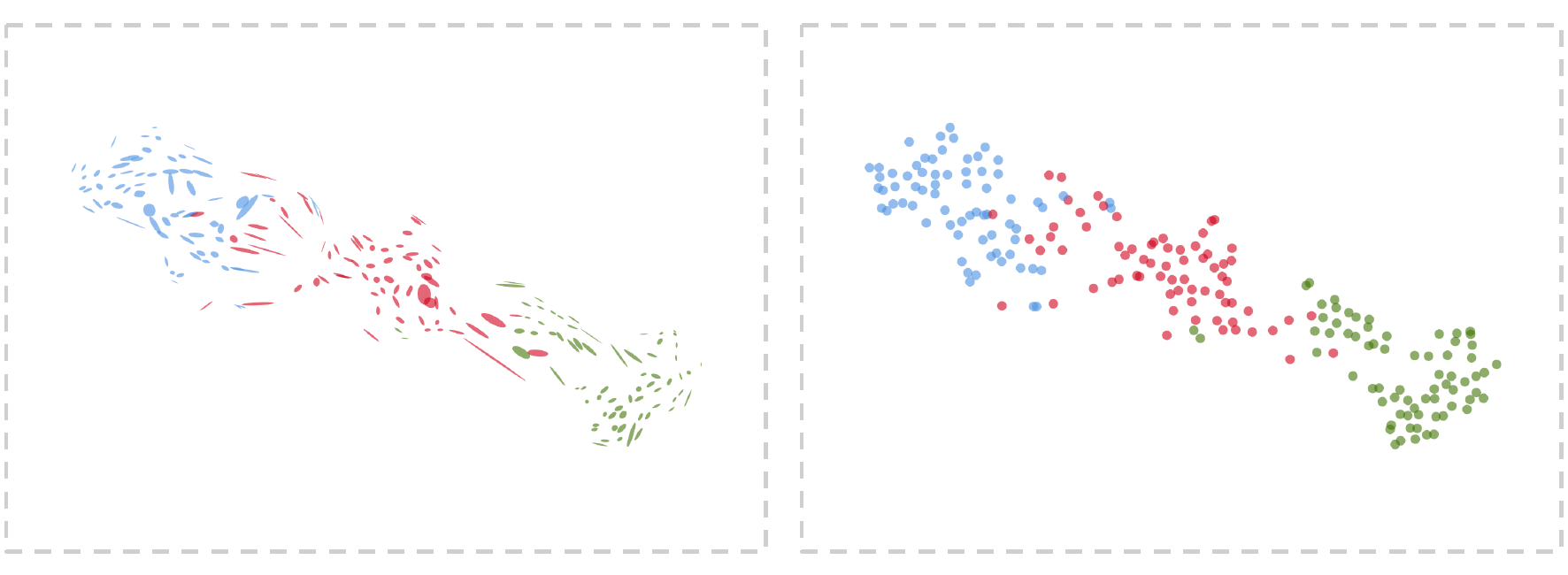}}\\	[-2ex]	
	\end{tabular}
	\caption{Visualizations of multidimensional datasets of Wine (first column) and Seeds (second column). Each projection method is visualized using our method (left boxes) and the point-based visualizations (right boxes). \rev{We can see global trends and twisting structures from the variations of local subspace glyphs. }}\vspace{-1em}
	\label{fig:multidim}
	\vspace{-3mm}
\end{figure*}


\begin{figure*}[!htb]
	\centering
	\begin{tabular}{cc}	
		\rot{90}{~~~~~~~~PCA}
		\subfloat{\includegraphics[width=0.95\linewidth]{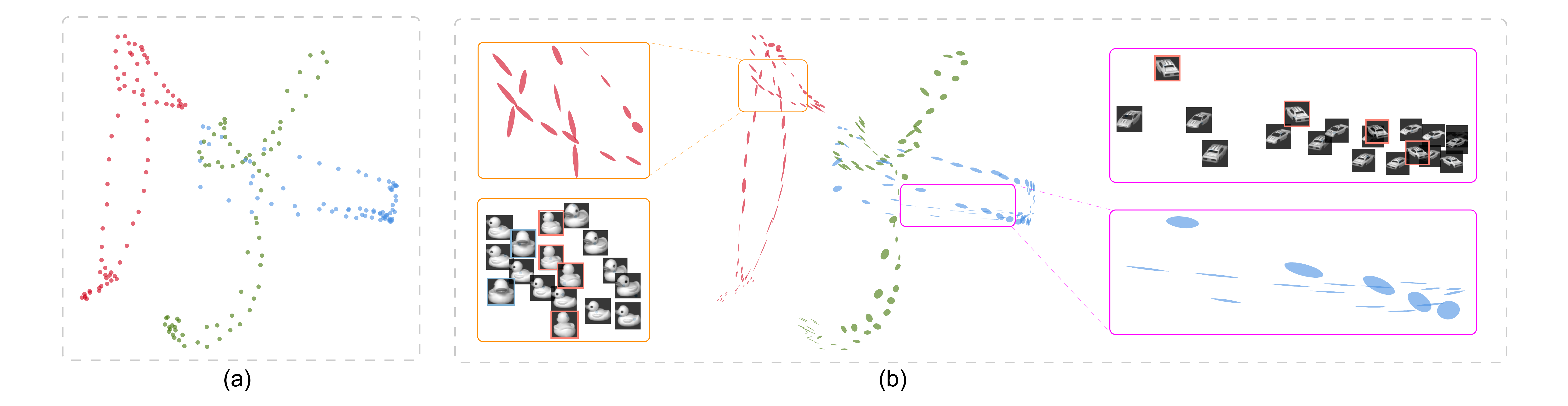}}\vspace{-1em}\\
		\rot{90}{~~~~~~~~MDS}
		\subfloat{\includegraphics[width=0.95\linewidth]{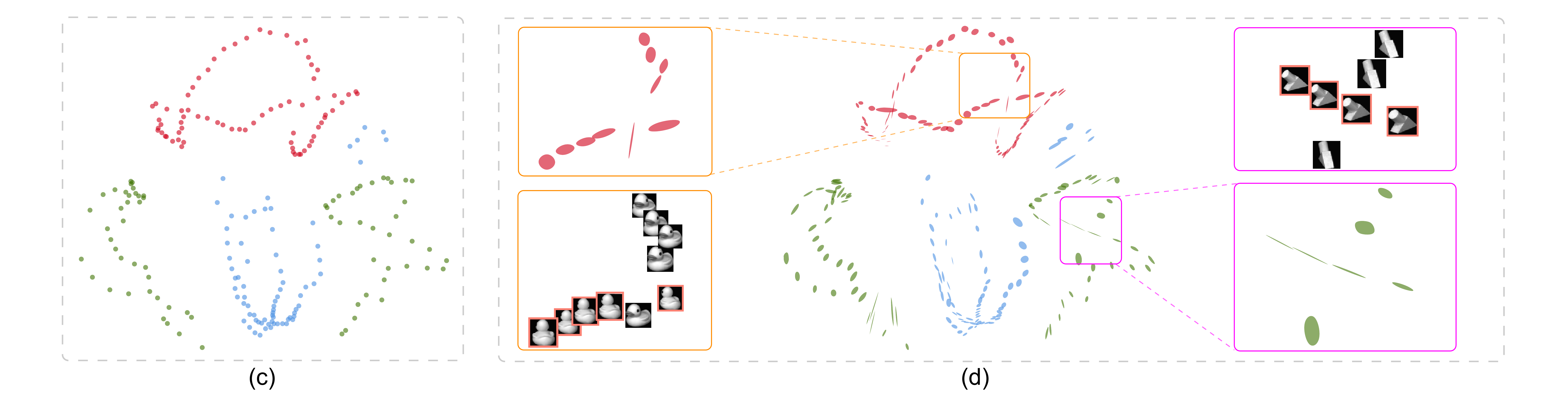}}\vspace{-1em}\\
		\rot{90}{~~~~~~~~t-SNE}
		\subfloat{\includegraphics[width=0.95\linewidth]{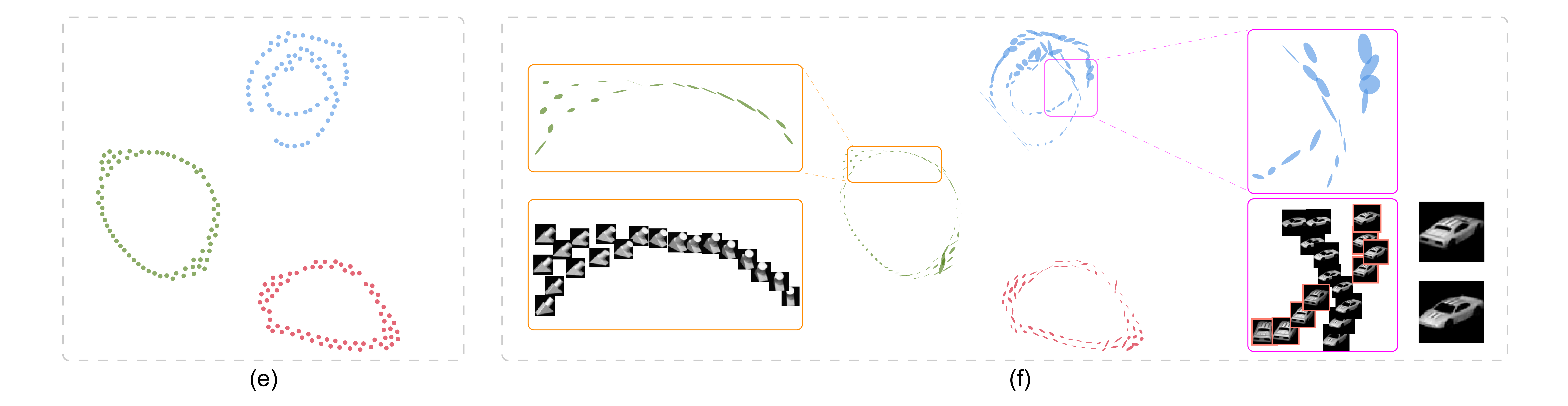}}\vspace{-1em}\\
	\end{tabular}
	\caption{Local subspace visualizations of the COIL dataset with PCA (first row), MDS (second row) shows that the different orientation and shape of the glyphs in adjacent areas correspond to different original data. And t-SNE (third row) shows that the orientation of glyphs changes smoothly within each cluster. }
	\label{fig:coil}
%
%
		\subfloat{\includegraphics[width=0.95\linewidth]{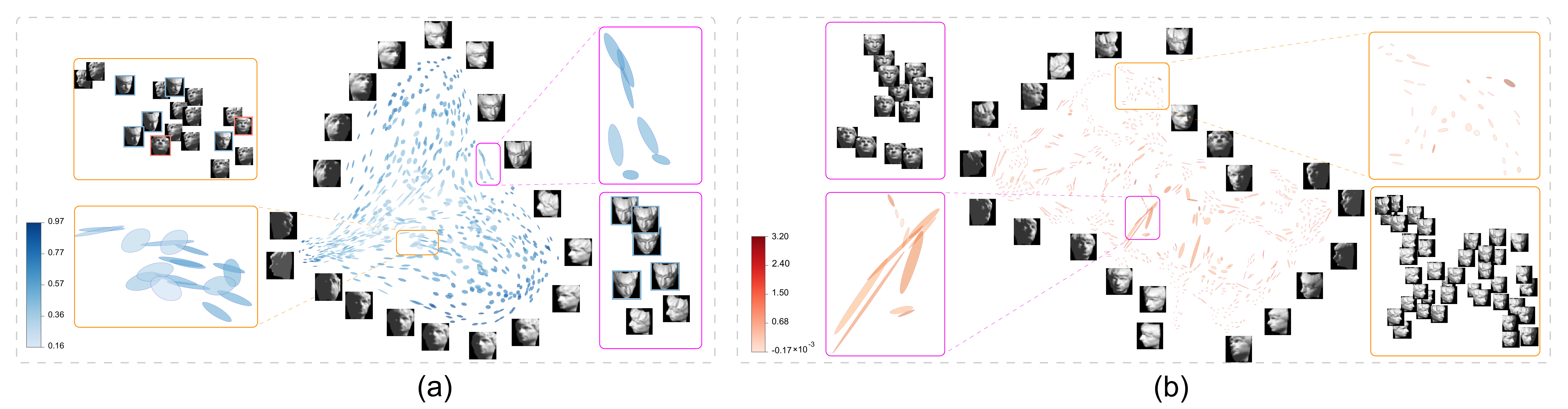}} \hfill
	\caption{Local subspace visualizations of the Face dataset with PCA (a) and t-SNE (b), which are colorzed by two different quality metrics:  trustworthiness and KL divergences, respectively. From the PCA result, we can see that two trends of glyphs separate around the center, one going upward and another going downward, and in each zoom-in, the distinct types of glyphs are associated with distinct face directions. From the t-SNE result, we can identify local patterns that exhibit trends. }
	\label{fig:face}
	\vspace{-5mm}
\end{figure*}

\section{Case Studies}
\label{sec:example}
In this section, we show examples of visual analysis of various representative multidimensional datasets---ranging from benchmark datasets in machine learning to downsampled image datasets in computer vision---to demonstrate the usefulness of our method.
We experiment with typical DR methods---PCA, MDS~\cite{borg2005modern}, and t-SNE~\cite{van2008visualizing}---on these datasets, and summarize and discuss interesting results.
Further examples can be found in the supplemental material.

Examples of multidimensional datasets from the UCI machine learning repository~\cite{Dua:2019} are shown in~\autoref{fig:multidim}.
The Wine dataset is a 13D data of 178 instances with 3 classes.
The PCA and MDS results as shown in~\autoref{fig:multidim} (Wine), rows one and two. With our method (left boxes), we can see global trends from the variations of local subspace glyphs.
In the PCA result, sizes of most glyphs are rather uniform, and the orientations of glyphs vary smoothly, forming a global arc structure; glyphs in the MDS projection suggest a swirling global structure.
Local subspace visualization of the t-SNE projection (\autoref{fig:multidim}, Wine, row three, left box) shows connections between the three clusters, and the glyphs on the exterior seem to form the boundary of the clusters---this may indicate that the clusters sit on the same manifold.
Besides the global trends, we can easily spot glyphs of distinct shapes (e.g., the elongated green glyphs in MDS and t-SNE) that can be further investigated.
With traditional point-based visualizations (right boxes), none of the global trends can be observed nor local points-of-interest can be found as indicated by abnormal glyphs.

Visualizations of the Seeds dataset (210 data points of 7D) do not exhibit clear global trends compared to the Wine example.
However, it can be observed that in the PCA (\autoref{fig:multidim}, Seeds, row one) and the MDS (\autoref{fig:multidim}, Seeds, row two) projections the glyphs form a twisting structure with crossings next to the central class.
The t-SNE result (\autoref{fig:multidim}, Seeds, row three) provides a similar look as of the Wine example; glyphs around the boundary of clusters seem to comprise the shape of a bow tie.

%


The COIL dataset~\cite{coil20} is a collection of 20 objects each of them photographed 72 times at different rotation angles.
We pick three objects: a duck, a car, and a block.
Scatterplots color-coded by object class visualizing projections of PCA, MDS, and t-SNE are shown in~\autoref{fig:coil}~(a), (c), and (e), respectively, where no correlation can be seen between points.
With our implicit local subspace projection method (\autoref{fig:coil}~(b)), global trends and local structures can be visualized with the orientation and shape of glyphs. In the orange zoom-in on the red class, we see that the transition of orientations of glyphs matches that of the ``duck" images. Four larger anisotropic glyphs are displayed in the purple zoom-in on the blue class---they are associated with backward facing ``cars." Furthermore, the anisotropy of glyphs is correlated with the rotation angle (with respect to front-facing images) of images.
In the MDS case (\autoref{fig:coil}~(row 2)), glyphs generated by our method (\autoref{fig:coil}~(d)) form twisting structures that are similar to the PCA case.
For example, by examining the images of the red glyphs in the orange zoom-in, it is confirmed that the crossings are actually associated to the change of orientation of the ``duck"; similarly, the green glyphs in the zoom-in on the right show two distinct directions of the ``block".
Our visualization of the t-SNE result (\autoref{fig:coil}~(f)) shows that the orientation of glyphs changes smoothly within each cluster. The orange zoom-in on the green cluster along with the actual images confirms that our method successfully captures the smoothly changing image orientations in the local subspaces. The purple zoom-in confirms that there exists a crossing of orientations of the car. 

The Face data contains 698 face images of different camera positions and lighting conditions.
Local subspace visualizations of the Face data with MDS is already shown in~\autoref{fig:teaser}, and PCA and t-SNE results are shown in~\autoref{fig:face}, respectively.
With PCA projection colorized by trustworthiness (\autoref{fig:face}~(a)), dense glyphs are seen on the left, and two trends of glyphs---one going upward and another going downward---separate around the center of the visualization. \yh{The intersection region between these two trends has lower trustworthiness and contains three distinct types of glyphs (see the zoom-in), corresponding to the face images with three distinct orientations. Similarly,  two distinct types of glyphs selected by the magenta box are associated with distinct face directions.}

\yh{The clusters can be clearly seen in our t-SNE result (\autoref{fig:face}~(b)), but no prominent global trends can be recognized. However, we can identify local patterns that exhibit trends; for example, in the orange zoom-in, the trends of glyphs are associated with the smooth change of face orientation.
By coloring each glpyh with the t-SNE loss function (KL divergence), we can see that the glyphs around the projection center has large size and loss values, indicating that the derivatives of the corresponding data points have large magnitudes.}

%

These experiments demonstrate the usefulness of our method in enhancing the anomaly identification ability and revealing global and local trends in data projections.

\section{Discussion, Conclusion, and Future Work}
We have introduced a method for visualizing local linear structures in multidimensional projections using implicit differentiation.
For each data sample in the original space, a local neighborhood is extracted and its structure is fitted using PCA.
Basis vectors of the multidimensional local neighborhood anchored at the respective data point are transformed to the low-dimensional visualization space by calculating differentiation of implicit functions. We have derived analytical solutions for typical multidimensional projections.
Next, for each data point, a glyph centered at the projected data point is generated by constructing a B-spline shape within the convex hull spanned by the transformed basis vectors.
To aid the analysis of our glyph representation, a linearity metric and a projection quality metric are used to depict the faithfulness of the transformation.
We have built an interactive, web-based visual analysis system that supports full-fledged, comparative analysis of our results and the point-based scatterplots.

We would like to discuss a few aspects that could affect the results of our method.
The number of neighbors $k$ of the local subspace is a user-set parameter---a small $k$ extracts local information of the data, whereas a large $k$ reveals more global information. We have observed that datasets with
rather high dimensionality (e.g., Face) are less sensitive to $k$ than datasets with lower dimensionality (e.g., Iris).
Our implicit differentiation transformation requires good convergence of the DR method---otherwise, numerical errors occur during the computation of the Jacobian matrix. Also, duplicate data points need to be removed as they would yield dividing by zero error when computing the Jacobian.

\rev{In terms of scalability, we argue that the visual scalability of our method is comparable to traditional point-based visualizations,	because the glyphs' appearance can be interactively adjusted, and most information can be perceived with point-sized (as in the point-based visualizations) glyphs as shown in examples throughout the paper.
	For larger datasets, our current method also suffers from the overdraw problem,
 as point-based visualizations in general, and we could adopt clutter reduction techniques~\cite{ellis2007taxonomy}, e.g., sampling, for better visual scalability.
	For computational scalability, the memory size is the limiting factor of computing the Jacobian matrix. The largest dataset we have experimented on our machine contains 7494 data points of 16-D. In the future, we would like to optimize the calculation of the Jacobian matrix on GPUs.
}

In conclusion, our method is a promising tool for understanding and analyzing multidimensional data and DR methods.
A benefit of our method is that main trends and anomalies in the projected data can be quickly identified with glyphs---large differences of orientation, shape, and size can be easily perceived.
Once the glyphs of interest are identified, the user can further analyze the data with interactive tools to understand the cause of trends or anomalies.
We have shown the usefulness of our method through a number of multidimensional datasets with popular DR methods, including PCA, MDS, and t-SNE.
In this process, we have gained new insights into familiar datasets and widely used DR methods.
Another benefit is that our implicit vector transformation is accurate and fast.
Our vector transformation method has been assessed through numerical comparisons, and our overall method has been evaluated through case studies on benchmark datasets.

\yh{In the future, we would like to integrate our approach with more DR methods together. One prominent example is UMAP~\cite{mcinnes2018umap}, which keeps a continuous transform function internally and thus can directly transform any point without our partial derivative based transformation. Second,
more intelligent analysis approaches could be used to assist analyzing our results, for example, pattern recognition methods that automatically detect trends and local patterns of interest.
Finally, we would like to study the effectiveness of our approach in real applications and compare with other explanation methods~\cite{faust2019dimreader}.
}



\acknowledgments{
This work is supported by the grants of the National Key Research \& Development Plan of China (2016YFB1001404), NSFC (61772315, 61861136012), by the Intel Graphics and Visualization Institutes of XeLLENCE, the National Institute of General Medical Sciences of the National Institutes of Health under grant numbers P41 GM103545 and R24 GM136986 and the Department of Energy under grant number DE-FE0031880 (USA), and by the Deutsche Forschungsgemeinschaft (DFG, German Research Foundation) Project-ID 251654672 -- TRR 161 (project B01).}
\bibliographystyle{abbrv-doi-hyperref}


\end{document}